\documentclass[11pt]{article}
\usepackage[final]{acl}
\usepackage{times}
\usepackage[T1]{fontenc}
\usepackage[utf8]{inputenc}
\usepackage{microtype}
\usepackage{inconsolata}
\usepackage{graphicx}
\usepackage{booktabs}
\usepackage{multirow}
\usepackage{adjustbox}
\usepackage{algorithm}
\usepackage[noend]{algpseudocode}
\algrenewcommand\algorithmiccomment[1]{\hfill{\footnotesize\color{black!60}$\triangleright$ #1}}
\algrenewcommand\alglinenumber[1]{{\footnotesize\color{black!50}#1}}
\usepackage{amsmath,amssymb,amsthm}
\newtheorem{proposition}{Proposition}
\theoremstyle{definition}
\newtheorem{lemma}{Lemma}\newtheorem{corollary}{Corollary}
\usepackage{xcolor,colortbl}
\usepackage{tikz}
\usepackage{tcolorbox}
\usepackage{pifont}
\usepackage{enumitem}
\setlist{nosep,leftmargin=*}
\definecolor{bccblue}{HTML}{00558A}
\definecolor{bccgreen}{HTML}{009E73}

\newcommand{\numTestAppworld}{168}
\newcommand{\numTestWebshop}{500}
\newcommand{\numTestAlfworld}{134}
\newcommand{\numAnatTwentysevenAwPfBZeroRuleHelp}{123}
\newcommand{\numAnatTwentysevenAwPfBZeroRuleHarm}{0}
\newcommand{\numAnatTwentysevenAwPfBZeroAnswerOn}{69.0}
\newcommand{\numAnatTwentysevenAwPfBZeroAnswerOff}{37.5}
\newcommand{\numAnatNineAwPfBFourRuleHelp}{14}
\newcommand{\numAnatNineAwPfBFourRuleHarm}{139}
\newcommand{\numAnatGemmaAwPfBThreeRuleHelp}{78}
\newcommand{\numAnatGemmaAwPfBThreeRuleHarm}{0}
\newcommand{\numAnatNineWsBTwoRuleHelp}{57}
\newcommand{\numAnatNineWsBTwoRuleHarm}{256}
\newcommand{\numVthreeAvgDone}{65.2}
\newcommand{\numVthreeFreeAvgDone}{63.8}
\newcommand{\numMainBestFullNineWs}{43.8}
\newcommand{\numMainRbcNineWs}{49.6}
\newcommand{\numMainBestFullGemmaAwPf}{65.3}
\newcommand{\numMainRbcGemmaAwPf}{69.8}
\newcommand{\numMainNoneTwentysevenAlf}{74.9}
\newcommand{\numMainFullTwentysevenAlf}{97.5}
\newcommand{\numMainCountersBelowNoneHarmful}{3}
\newcommand{\numKAvgFull}{51.2}
\newcommand{\numKAvgCut}{61.9}
\newcommand{\numKAvgRbcOne}{63.1}
\newcommand{\numKAvgBestFullOne}{62.2}
\newcommand{\numKAvgRbcTwo}{65.7}
\newcommand{\numKAvgRbcThree}{67.2}
\newcommand{\numKAvgBestFullThree}{66.1}
\newcommand{\numKDropRbcThree}{0.8}
\newcommand{\numKDropFull}{69.6}
\newcommand{\numKPickWithinHalf}{7}
\newcommand{\numSysDcNone}{53.8}
\newcommand{\numSysDcFull}{61.3}
\newcommand{\numSysDcCut}{61.1}
\newcommand{\numSysDcBestfull}{66.9}
\newcommand{\numSysDcRbc}{66.5}
\newcommand{\numSysAllNone}{54.1}
\newcommand{\numSysAllFull}{55.5}
\newcommand{\numSysAllRbc}{66.9}
\newcommand{\numSysDcSettings}{8}
\newcommand{\numSysAllSettings}{19}
\newcommand{\numSysDcSpread}{37.3}
\newcommand{\numSysDcSpreadWhere}{Gemma ALFWorld}
\newcommand{\numSysDcPicksWhole}{7}
\newcommand{\numSysCrossR}{50}
\newcommand{\numCostScoreAppWorld}{171}
\newcommand{\numCostRatioMin}{3.3}
\newcommand{\numCostRatioMax}{4.1}
\newcommand{\numCostRatioBestFullMin}{1.8}
\newcommand{\numCostRatioBestFullMax}{2.6}
\newcommand{\numKRho}{0.89}
\newcommand{\numKPickBest}{6}
\newcommand{\numKPickN}{11}
\newcommand{\numKWideNone}{78.4}
\newcommand{\numTagsRho}{0.21}
\newcommand{\numTagsBlocks}{81}
\newcommand{\numRunsTwentysevenAwPtOne}{82.5}
\newcommand{\numRunsTwentysevenAwPtTwo}{58.9}
\newcommand{\numRunsTwentysevenAwPtThree}{30.0}
\newcommand{\numRunsHarmSettings}{6}
\newcommand{\numRunsSwing}{3}
\newcommand{\numRunsHarmPts}{5}
\newcommand{\numRunsFirstHelps}{5}
\newcommand{\numRunsFirstLevel}{2}
\newcommand{\numRunsFirstHarms}{4}
\newcommand{\numReevalMax}{5.0}
\newcommand{\numRunsDcNineWsGain}{13.2}
\newcommand{\numRunsDcNineWsLoss}{14.0}
\newcommand{\numLgTwentysevenAwPtNone}{78.4}
\newcommand{\numLgTwentysevenAwPtFull}{46.6}
\newcommand{\numLgTwentysevenAwPtBf}{78.4}
\newcommand{\numLgTwentysevenAwPtRbc}{84.9}
\newcommand{\numLgNineAwPtNone}{19.8}
\newcommand{\numLgNineAwPtFull}{17.7}
\newcommand{\numLgNineAwPtBf}{21.2}
\newcommand{\numLgNineAwPtRbc}{32.3}
\newcommand{\numLgMeanGain}{3.1}
\newcommand{\numLgCutN}{6}
\newcommand{\numLgCutTokMax}{44}
\newcommand{\numLgCutGain}{4.7}
\newcommand{\numAkBestFullFive}{66.8}
\newcommand{\numAkCurationPct}{17}
\newcommand{\numAkServePct}{27}
\newcommand{\numAkServePctMatched}{23}
\newcommand{\numSysAboveFullAll}{15}
\newcommand{\numSelNonDevGain}{+1.9}
\newcommand{\numSelPicksDiffer}{6}
\newcommand{\numSelAhead}{5}
\newcommand{\numCrossDcGain}{2.2}
\newcommand{\numCrossAceGain}{0.4}
\newcommand{\numMainRuleDropMin}{60.3}
\newcommand{\numMainRuleAvgMax}{50.9}
\newcommand{\numMainRuleAvgMaxDc}{58.9}
\newcommand{\numMainOwnDropAce}{72.6}
\newcommand{\numMainOwnAvgAll}{53.8}
\newcommand{\numMainBfTopkAce}{62.6}
\newcommand{\numMainBfTopkDc}{62.3}
\newcommand{\numMainBaselineDropMin}{5.4}
\newcommand{\numMainTokRbcAce}{60}
\newcommand{\numMainTokBfAce}{78}

\newcommand{\bcc}{RBC}

\newcommand{\mNone}{No playbook}
\newcommand{\mFull}{Whole}
\newcommand{\mRandom}{Random}
\newcommand{\mCounters}{Tags}
\newcommand{\mOwn}{Own}
\newcommand{\mTopk}{Top-$k$}

\newcommand{\mDCo}{DC\textsubscript{outcome}}
\newcommand{\mDCs}{DC\textsubscript{self}}

\newcommand{\mCountersT}{Tags$^{*}$}
\newcommand{\mTopkT}{Top-$k^{*}$}
\newcommand{\mCut}{\bcc{}-cut}
\newcommand{\mBestFull}{Best-Full}
\newcommand{\Dpr}{\mathcal{D}_{\mathrm{pr}}}

\newcommand{\yes}{$\bullet$}
\newcommand{\half}{$\circ\!\!\bullet$}
\newcommand{\no}{$\circ$}
\newcommand{\tpar}[1]{\par\vspace{3pt}\noindent\textbf{#1}\hspace{0.5em}\ignorespaces}
\newenvironment{appproof}[1][Proof]{\par\addvspace{4pt}\setlength{\parindent}{0pt}\setlength{\parskip}{3pt}\noindent\textbf{#1.}\hspace{0.5em}\ignorespaces}{\par\addvspace{4pt}}
\newdimen\fitextra
\newcommand{\fittable}[2]{%
  \setbox0=\hbox{#2}%
  \fitextra=\linewidth \advance\fitextra by -\wd0 \divide\fitextra by #1\relax
  \advance\tabcolsep by \fitextra
  \ifdim\tabcolsep<3pt \tabcolsep=3pt\fi
  \setbox0=\hbox{#2}%
  \ifdim\wd0>\dimexpr\linewidth+0.1pt\relax
    \PackageInfo{bcc}{a table is wider than its column (\the\wd0\space vs \the\linewidth); scaled to fit}%
    \noindent\resizebox{\linewidth}{!}{\box0}%
  \else
    \noindent\makebox[\linewidth][c]{\box0}%
  \fi}
\newcommand{\hb}[1]{{\bfseries\boldmath #1}}
\newcommand{\hd}[1]{\multicolumn{1}{c}{\bfseries\boldmath #1}}
\newcommand{\hdR}[1]{\multicolumn{1}{c@{}}{\bfseries\boldmath #1}}
\newcommand{\hs}[1]{\multicolumn{1}{c}{#1}}
\newcommand{\hsR}[1]{\multicolumn{1}{c@{}}{#1}}
\newcommand{\hg}[2]{\multicolumn{#1}{c}{\bfseries\boldmath #2}}
\newcommand{\hbm}[1]{\multirow{2}{*}{\bfseries\boldmath #1}}
\newcommand{\hdm}[1]{\multicolumn{1}{c}{\multirow{2}{*}{\bfseries\boldmath #1}}}
\newcommand{\hdmR}[1]{\multicolumn{1}{c@{}}{\multirow{2}{*}{\bfseries\boldmath #1}}}
\newtcolorbox{lessonbox}{colback=black!8, colframe=black!30, boxrule=0.4pt, arc=2.5pt, left=7pt, right=7pt,
  top=8pt, bottom=8pt, before skip=11pt, after skip=11pt, fontupper=\small\itshape}
\newtcolorbox{promptbox}[1]{colback=black!4, colframe=black!30, colbacktitle=black!12, coltitle=black, boxrule=0.4pt,
  arc=2pt, left=4pt, right=4pt, top=3pt, bottom=3pt, toptitle=1pt, bottomtitle=1pt, before skip=6pt, after skip=6pt,
  fonttitle=\footnotesize\bfseries, fontupper=\footnotesize, title={#1}}

\title{Reliable Memory from Unstable Growth: Curating Self-Written Agent Playbooks with Randomized Experiments}
\author{
  \textbf{Hoang Pham}\textsuperscript{1} \quad
  \textbf{Cong-Duy Nguyen}\textsuperscript{2} \quad
  \textbf{Dong Le}\textsuperscript{1} \\
  \textbf{Ponhvoan Srey}\textsuperscript{1} \quad
  \textbf{Anh Tuan Luu}\textsuperscript{1,2}\thanks{Corresponding author.} \\
  \textsuperscript{1}Nanyang Technological University \quad
  \textsuperscript{2}VinUniversity
  \vspace{0.2cm}\\
  \texttt{Email:\hspace{0.4em}pham0100@ntu.edu.sg}
}

\begin{document}
\maketitle
\begin{abstract}
Persistent textual memory lets language-model agents reuse lessons from past tasks to improve on future multi-step tasks without retraining.
Systems such as ACE and Dynamic Cheatsheet keep it as a \emph{playbook}: lessons the agent writes after each task and reads in its prompt.
A key challenge is to decide which lessons improve later tasks.
Current systems leave the decision to the agent itself, which makes performance unstable: the same procedure can write a playbook that helps or one that falls up to \numKDropFull{} points below no playbook.
These findings motivate valuing lessons by their measured effect on task success rather than by the agent's judgement.
We propose \textbf{Randomized Block Curation (\bcc)}\footnote{Our code is available at \url{https://anonymous.4open.science/r/rbc}.}, which measures this effect with a randomized experiment on practice tasks.
It groups related lessons into blocks large enough to move success measurably and, on each practice task, includes or leaves out every block at random.
Random inclusion makes tasks with and without a block comparable, so the gap in their success estimates the block's effect.
\bcc{} keeps the blocks that help and, since the same procedure can write a helpful or a harmful playbook, grows several and serves whichever curated, whole or no playbook scores best on fresh practice episodes.
Across two memory systems, three models and three benchmarks, \bcc{} raises average success to \numSysAllRbc, against \numSysAllFull{} with the whole playbook, \numSysAllNone{} with no playbook and \numMainOwnAvgAll{} when the agent's own judgement picks the lessons to keep.
\end{abstract}

\section{Introduction}
\label{sec:intro}

\begin{figure}[t]
\centering
\includegraphics{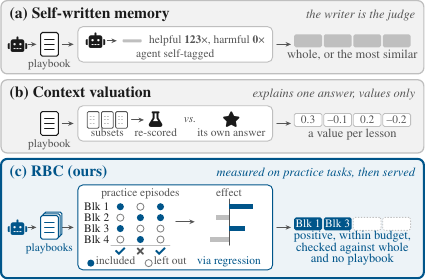}
\caption{\textbf{Judging versus measuring a playbook.} \textbf{(a)} The agent judges its lessons and serves the playbook whole or by similarity. \textbf{(b)} Context valuation
re-scores random subsets to explain one answer, not to serve. \textbf{(c)} \bcc{} includes each block at random on practice tasks, keeps those that help within a budget, and serves them unless a whole or no playbook
scores higher.}
\label{fig:approaches}
\end{figure}

Language-model agents carry out multi-step tasks such as operating apps, shopping online and household chores
\citep{trivedi2024appworld, yao2022webshop, shridhar2021alfworld}. Without retraining, which is costly
or impossible through a proprietary API, a growing line of work lets agents improve through persistent textual memory: lessons
the agent writes after each task and reads in its prompt later \citep{shinn2023reflexion, zhao2024expel},
which we call a \emph{playbook}. Systems maintain it differently, growing it lesson by lesson
\citep[ACE;][]{zhang2025ace}, rewriting it as a whole \citep[Dynamic Cheatsheet;][]{suzgun2025dynamic} or retrieving
lessons by similarity \citep[ReasoningBank;][]{ouyang2025reasoningbank}, but they share one design
(Figure~\ref{fig:approaches}a): the agent that writes the lessons also judges which to keep.

Few works test this judgement against task success, and those that do compare whole memories \citep{rsea2026}. This
matters because added context can distract a model as easily as help it \citep{shi2023distracted}, and a lesson kept by
mistake is served on every later task. This work targets that decision, which lessons of a self-written playbook to serve, in two established memory systems, ACE and Dynamic Cheatsheet. We find their playbooks unstable: served whole, a playbook can fall up to \numKDropFull{} points below no playbook, and growing it again on the same tasks in another order can turn a helpful playbook harmful. Whether a playbook helps depends on the growth run that wrote it and is known only by measuring its effect on success.

Letting the agent prune lessons it judges harmful does not remove the harm. Judging a lesson requires comparing episodes with and without it, but the agent sees only those that use it, and language
models are poor and biased judges of their own outputs \citep{huang2024selfcorrect, zheng2023judging}. So it tags lessons as helpful when they sound right, not when they raise success, even a harmful rule (Section~\ref{sec:analysis}). Context-valuation methods run this comparison but not for curation: they explain why a model gave one answer \citep{cohenwang2024contextcite} or score memory entries one at a time \citep{skillsv2026, dgmem2026}, though one entry rarely moves success measurably, and return scores, not a playbook to serve (Figure~\ref{fig:approaches}b).

We propose \textbf{Randomized Block Curation (\bcc)}, which supplies this comparison with a randomized experiment on
practice tasks (Figure~\ref{fig:approaches}c). \bcc{} groups related lessons into blocks large enough to move success measurably. Including each block at random on every practice task makes tasks with and without it comparable, so their success gap estimates its effect. \bcc{} keeps the blocks that help within a token budget and, since a playbook's value depends on its run, grows several and serves whichever curated, whole or no playbook scores best on fresh practice episodes. \bcc{} leaves growth itself unchanged.

We test \bcc{} on both systems with three models on three benchmarks. It lifts both on every model, averaging
\numSysAllRbc{} task success against \numSysAllFull{} with the whole playbook, \numSysAllNone{} with no playbook and
\numMainOwnAvgAll{} when the agent picks lessons itself. It also avoids the harm: on ACE it never falls more than
\numKDropRbcThree{} points below no playbook, while every baseline falls at least \numMainBaselineDropMin{} points
below it in some setting. When ACE's playbooks, which never drop a lesson, grow for twice as long, \bcc{} still beats the whole playbook and no playbook in every setting.

\noindent In summary, our contributions are as follows.
\begin{itemize}
  \item \textbf{Analysis.} We show that self-written playbooks are unstable across growth runs and that the agent's own judgement misses the harm.
  \item \textbf{Method.} We propose \bcc{}, which measures each block of lessons in a randomized experiment and serves only what helps.
  \item \textbf{Experiments.} \bcc{} is effective and robust across memory systems, models, benchmarks and rounds of memory updates.
\end{itemize}

\section{Related Work}
\label{sec:related}
\noindent\textbf{Self-written agent memory.}\hspace{0.5em}Of the systems above, ACE tags each lesson as
helpful or harmful, and Dynamic Cheatsheet counts each entry's successful uses. Other agents reflect on failures or store workflows
\citep{shinn2023reflexion, wang2025awm}, memory managers let the model decide what to store and recall
\citep{packer2023memgpt, xu2025amem}, and ExpeL \citep{zhao2024expel} scores each insight by the model's votes.
In all of them a lesson stays because its writer believes it helps or because it resembles the task;
\citet{seal2026} find that self-improving agents also misjudge the checks they write, and we find the same for the lessons agents keep. The closest works measure success, but on whole units: \citet{sarukkai2025selfgen} select among sets of trajectories and RSEA \citep{rsea2026} among memory rewrites, accepting one only if held-out success does not drop. \bcc{} instead measures every block at once and chooses among growth runs and no playbook.

\tpar{Valuing parts of a context.}ContextCite \citep{cohenwang2024contextcite} re-scores one answer on random subsets of sources with a linear fit, the design \bcc{} applies to task success; Data Shapley \citep{ghorbani2019data} and Data Banzhaf \citep{wang2023databanzhaf} value
training data, and SkillSV \citep{skillsv2026} and DG-Mem \citep{dgmem2026} value agent skills or memory entries.
They stop at a value per entry, and a single entry rarely moves success beyond the noise of a few hundred episodes
\citep{coalitionaware2026}; \bcc{} measures blocks, scores them by task success and decides what to serve within
a budget (Appendix~\ref{app:related}). A block's effect, its average gain, is its Banzhaf value \citep{banzhaf1965weighted}, which random inclusion estimates for every block at once.

\tpar{Choosing prompts and harmful context.}Prompt optimizers such as DSPy \citep{khattab2024dspy} and GEPA
\citep{agrawal2026gepa} keep what scores best on a development set, and demonstration selection scores each
example over random prompts that contain it \citep{chang2023data, nguyen2023incontext}; \bcc{} applies this scoring to an agent's memory, selecting among the agent's lessons instead of rewriting them, so curating calls no language model (Appendix~\ref{app:prompts}). Irrelevant text distracts language models \citep{shi2023distracted, liu2024lost}, so a longer playbook can hurt, and prompt compression \citep{jiang2023llmlingua} cuts a context by perplexity, not by effect on success; run-to-run variance is familiar from fine-tuning \citep{dodge2020finetuning}.

\section{Background and Setup}
\label{sec:background}

\subsection{Memory systems}
\label{sec:memory}
\tpar{The memory loop.}A self-written memory system lets an agent learn from its own experience. It wraps three
steps around the agent:
\begin{enumerate}
  \item \textbf{Grow.} After each batch of training tasks, the model reads its own trajectories and writes short
        lessons, which form a \emph{playbook} $P$.
  \item \textbf{Judge.} The model judges which lessons help.
  \item \textbf{Serve.} A subset $S \subseteq P$ of the lessons goes into the prompt of later tasks.
\end{enumerate}
On a task $x$, the agent acts over many turns, an \emph{episode}, and earns a success score
$y(x, S) \in [0, 1]$ when $S$ is served (reported $\times 100$). The \emph{value} of serving $S$ is its mean success,
$V(S) = \mathbb{E}[\, y(x, S) \,]$, and a playbook helps only when $V(S) > V(\emptyset)$. Our two
\emph{reference points} are \mNone{} ($S = \emptyset$) and the whole playbook (\mFull, $S = P$). A \emph{growth
run} grows one playbook from scratch on the training tasks. Runs differ: with the tasks in another order and
new sampling, the same procedure writes a different playbook, one that may help or harm.

\tpar{Goal.}In this work, we replace the judge and serve steps by measurement. For a playbook $P$ and a token budget $B$, the best set of lessons to serve is
\begin{equation}
  \textstyle S^{\star} \in \arg\max_{S \subseteq P,\; \mathrm{tokens}(S) \le B} \, V(S) ,
  \label{eq:problem}
\end{equation}
and with several growth runs, the best of their sets $S^{\star}$, whole playbooks and no playbook. As $V$ is seen only in noisy episodes, we estimate it on practice tasks (Section~\ref{sec:setup}) before serving.

\tpar{Two systems.}We study two established systems that write a playbook in different ways, to test whether a
finding depends on how memory is written; ReasoningBank's retrieval by similarity serves as a baseline (Section~\ref{sec:experiments}).

\tpar{ACE.}Our main system, ACE~\citep{zhang2025ace}, grows its playbook lesson by lesson. As the agent solves
training tasks, a \emph{Reflector} reads each trajectory and its feedback and writes new lessons; no lesson is ever
removed. The Reflector also judges: it tags the lessons used in each episode as helpful or harmful, a record we can
check (Appendix~\ref{app:ace}). ACE serves the whole playbook.

\tpar{Dynamic Cheatsheet.}Our second system \citep{suzgun2025dynamic} rewrites its whole playbook, a cheatsheet:
after each batch of training tasks, one model call reads it and the batch's trajectories and writes a
new one. It judges by itself, without the reward, which attempts succeeded, counts each entry's
successful uses (Appendix~\ref{app:dc}) and, like ACE, serves its whole playbook.

\subsection{Benchmarks and models}
\label{sec:setup}
\tpar{Benchmarks.}We conduct experiments on three benchmarks. In AppWorld~\citep{trivedi2024appworld}, the agent
writes code against the APIs of nine apps. In WebShop~\citep{yao2022webshop}, it finds and buys a described
product. In ALFWorld~\citep{shridhar2021alfworld}, it reaches a goal in a text household. They are scored by
task goal completion, purchase reward and success rate.

\tpar{Task splits.}Each benchmark has three disjoint task sets: \emph{training tasks} to grow playbooks,
\emph{practice tasks} to test playbooks before serving one, and \emph{evaluation tasks} for the final score: AppWorld \numTestAppworld, WebShop \numTestWebshop{} and ALFWorld \numTestAlfworld{} (Table~\ref{tab:cells}).

\tpar{Models and settings.}We run three open models: \mbox{Qwen3.8-27B} \citep{qwen2026qwen38}, Qwen3.5-9B
\citep{qwen2026qwen35} and Gemma-4-31B \citep{gemma2026gemma4}. A \emph{setting} is one system, model and benchmark. Per model, ACE has four, two of them on AppWorld, where its Reflector sees whether a task passed (\emph{pass/fail}) or each unit test's result (\emph{per-test}); Dynamic Cheatsheet, which never sees the reward, has three. Without 9B on ALFWorld (it solves almost no game), that makes eleven ACE and eight Dynamic Cheatsheet settings. Each setting has three growth runs.

\section{Self-Written Memory Is Unreliable}
\label{sec:analysis}

\noindent Before repairing self-written playbooks, we ask whether they need repair. We ran three growth runs in
each of the eleven ACE settings and checked each step of the loop (serve, grow, judge) against task success. We call a playbook
\emph{harmful} when it scores more than \numRunsHarmPts{} points below no playbook; re-evaluating a playbook moved its score by at most \numReevalMax{} points, so such a gap is not noise.

\begin{figure}[!h]
\centering
\includegraphics[width=\columnwidth]{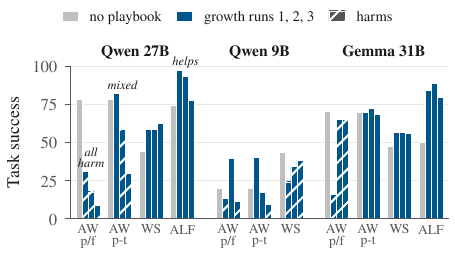}
\caption{The same procedure grows helpful and harmful playbooks: three ACE growth runs per setting. Hatched: more than \numRunsHarmPts{} points below no playbook. AW: AppWorld with pass/fail (p/f) or per-test (p-t) feedback; WS: WebShop; ALF: ALFWorld.}
\label{fig:runs}
\end{figure}

\tpar{Finding 1: serving the whole playbook can harm.}Served whole, a playbook is unstable, even on one model and in one setting. It can lift success a long way: on 27B ALFWorld (Figure~\ref{fig:runs}, \emph{helps}) the first playbook raises it to \numMainFullTwentysevenAlf{} from
\numMainNoneTwentysevenAlf{} with no playbook. It can also erase most of it: on 27B AppWorld with pass/fail feedback (\emph{all harm}) every playbook harms, one by \numKDropFull{} points. Of the first growth runs in the eleven settings, \numRunsFirstHelps{} help,
\numRunsFirstLevel{} are level and \numRunsFirstHarms{} harm. A playbook is extra context, and context that does not
help can distract a model \citep{shi2023distracted, liu2024lost}, so serving one is a decision, not a default.

\tpar{Finding 2: the growth run decides whether the playbook harms.}If this harm came from the model or benchmark, regrowing the playbook would reproduce it. It does not always: the same procedure on the same tasks, in another order, can write a helpful or a harmful playbook. On 27B AppWorld with per-test feedback,
the three runs score \numRunsTwentysevenAwPtOne, \numRunsTwentysevenAwPtTwo{} and \numRunsTwentysevenAwPtThree{}
against \numKWideNone{} with no playbook (Figure~\ref{fig:runs}, \emph{mixed}): one helps, two harm. Such swings occur in \numRunsSwing{} of the
\numRunsHarmSettings{} settings with a harmful run, all on Qwen, whose runs differ only in task order and sampling. Dynamic Cheatsheet shows them too (Appendix~\ref{app:dc-runs}): on 9B WebShop one run gains \numRunsDcNineWsGain{} points over no playbook, another loses \numRunsDcNineWsLoss. An agent with one playbook cannot tell which kind it holds; only measuring can.

\begin{figure}[!h]
\centering
\includegraphics[width=\columnwidth]{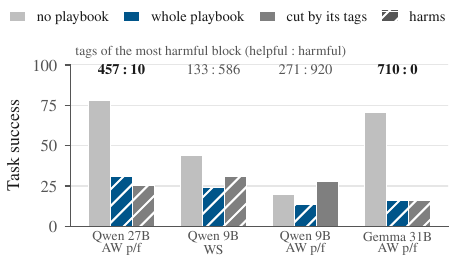}
\caption{Cutting by the agent's own tags (to 40\% of the tokens) leaves the harm in three of the four settings where the first ACE playbook harms. Above the bars: tags of the block that lowers success most, bold where they call it helpful. Hatched: more than \numRunsHarmPts{} points below no playbook. AW p/f: AppWorld, pass/fail feedback; WS: WebShop.}
\label{fig:tags}
\end{figure}

\tpar{Finding 3: the agent's own judgement misses the harm.}The obvious defence is to let the agent's tags prune the playbook: keep the lessons tagged most helpful and drop the rest. Yet in \numMainCountersBelowNoneHarmful{} of the
\numRunsFirstHarms{} settings where the first playbook harms, the cut playbook still harms
(Figure~\ref{fig:tags}). The tags are self-judgement, and language models are poor and biased
judges of their own outputs \citep{huang2024selfcorrect, zheng2023judging}: the Reflector sees a lesson followed, not
what it costs. On AppWorld, the most harmful block
holds rules like this one from the 27B playbook grown with pass/fail feedback:
\begin{lessonbox}\raggedright
``When calling \texttt{complete\_task}, always pass a concrete, non-empty string literal as the answer argument.''
\end{lessonbox}
\noindent The rule sounds careful, but many tasks expect no answer: with such rules in the prompt, 27B gives one on
\numAnatTwentysevenAwPfBZeroAnswerOn\% of these tasks, against \numAnatTwentysevenAwPfBZeroAnswerOff\% without them.
Every task ends with this call, so the rule bears on every episode and collects tags in all of them: the Reflector
tagged it helpful \numAnatTwentysevenAwPfBZeroRuleHelp{} times and never harmful. Shown the failed tests under per-test feedback, it writes no such rule. On the two larger models, the most harmful block carries almost only
helpful tags (above the bars in Figure~\ref{fig:tags}): the agent's judgement points the wrong way. Such harm shows
only in task success, so we measure it as \bcc{} does, comparing success with and without each block, served at
random on practice tasks.

\begin{figure*}[t]
\centering
\includegraphics[width=0.95\textwidth]{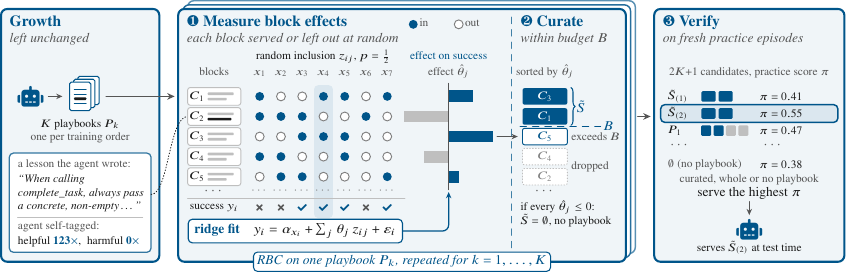}
\caption{Overview of \bcc. Growth is unchanged: the agent writes $K$ playbooks and tags its own lessons (the quoted
lesson is real; its block lowers success although tagged helpful). 1)~Each block is included or left out at random in
every practice episode, and a regression of success on these choices estimates its effect $\hat\theta_j$. 2)~Each
playbook is curated to the blocks with a positive effect, best first, within the budget $B$. 3)~Every candidate is
scored on fresh practice episodes and the best is served. Other values are illustrative.}
\label{fig:main}
\end{figure*}

\tpar{What a repair needs.}Each finding sets one requirement: serve only the parts of a playbook that help, or no
playbook (Finding~1); grow several playbooks and let fresh practice episodes choose among them (Finding~2); and
measure each part's effect on task success (Finding~3).

\section{Method: Randomized Block Curation}
\label{sec:method}

\noindent In this section, we introduce \bcc{}, which replaces the judge and serve steps of the memory loop with
measurement (Figure~\ref{fig:main}; pseudocode in Appendix~\ref{app:pseudocode}). It runs after growth, on practice
tasks $\Dpr$ kept apart from the evaluation tasks. It needs only a list of lessons, so it works on either system's
playbook: ACE's lessons or the entries of Dynamic Cheatsheet's cheatsheet. It meets the three requirements above in
three steps: it measures the effect of each block of lessons, curates each grown playbook by these effects, and
verifies the curated playbooks against the whole playbooks and no playbook on fresh practice episodes. Each curated
set $\tilde S$ estimates the best set $S^{\star}$ of Eq.~\eqref{eq:problem} from the noisy outcomes of practice episodes alone.

\subsection{Measuring the effect of each block}
\label{sec:pricing}
\tpar{Blocks.}We measure groups of related lessons, as a single lesson rarely moves success beyond
the noise of a few hundred episodes. We embed each lesson with all-mpnet-base-v2 \citep{reimers2019sentencebert}, as ACE does to find duplicates, and cluster the embeddings with Ward's method \citep{ward1963hierarchical} into \mbox{$m = 8$} \emph{blocks} $C_1, \dots, C_m$ (the number
matters little; Appendix~\ref{app:sensitivity}). A block holds lessons on one topic, which tend to help or harm together. For a set of blocks $T \subseteq [m]$, we write $V(T)$ for the value of serving their lessons.

\tpar{Randomized inclusion.}Each practice episode $i$ draws a task $x_i$ and includes each block $j$ independently with probability $p = \frac12$, \mbox{$z_{ij} \sim \mathrm{Bernoulli}(p)$}; the agent is served the included blocks, $T_i = \{\, j : z_{ij} = 1 \,\}$, and scores $y_i$. Each episode is thus both an ordinary run of the agent and
one trial of a two-level factorial design over all $m$ blocks, each block on or off (\citealp{box2005statistics};
Figure~\ref{fig:main}, step~1).

\tpar{What is estimated.}A block's gain can depend on the other blocks served with it, so we define its effect as its average gain over them, which at $p = \frac12$ form a uniformly random subset $T \subseteq [m] \setminus \{j\}$:
\begin{equation}
  \textstyle \theta_j \;=\; \mathbb{E}_{T} \big[\, V(T \cup \{j\}) - V(T) \,\big] ,
  \label{eq:estimand}
\end{equation}
the Banzhaf value of block $j$ \citep{banzhaf1965weighted}: how much the block raises success, averaged over what else is served with it. Random inclusion makes it measurable by a plain comparison of the episodes with and without the block:
\begin{proposition}
\label{prop:estimate}
If every block is included independently with probability $\frac12$, then for every block $j$ and any $V$,
$\mathbb{E}[y \mid z_j = 1] - \mathbb{E}[y \mid z_j = 0] = \theta_j$. Over $n$ episodes, this difference of means, taken within tasks, is unbiased for $\theta_j$, for all $m$ blocks at once, with variance about $4\sigma^2/n$, where $\sigma^2$ bounds
the within-task variance of $y$.
\end{proposition}
The measurement thus needs no assumption on $V$ and few episodes: interactions between blocks are averaged into
$\theta_j$, not assumed away (a block that adds $g$ only alongside another gets $\theta_j = g/2$), and one round of
episodes measures every block, where leaving out one block at a time would need \mbox{$(m+1)/2 = 4.5$} times as many
episodes for the same precision. We compute $\hat\theta$ by a ridge regression of success on the inclusions, with one
intercept per task to absorb task difficulty, minimizing
\begin{equation}
  \textstyle \sum_i \big( y_i - \alpha_{x_i} - z_i^{\top}\theta \big)^2 + \lambda \lVert \theta \rVert^2 ,
  \label{eq:ridge}
\end{equation}
with the task intercepts $\alpha$ unpenalized and $\lambda$ chosen by leaving one task out. Unpenalized, this fit is
consistent for $\theta_j$; under this balanced design, the penalty scales all estimates by about the same factor,
keeping signs and order (Appendix~\ref{app:banzhaf}).

\subsection{Curating under a budget}
\label{sec:curation}
The curated set uses only the signs and order of $\hat\theta$ (Figure~\ref{fig:main}, step~2):
\begin{enumerate}
  \item keep the blocks with $\hat\theta_j > 0$, highest $\hat\theta_j$ first;
  \item add their lessons in that order, skipping any lesson that would exceed $B$; the result is $\tilde S$;
  \item if no estimate is positive, $\tilde S = \emptyset$ (no playbook).
\end{enumerate}
In Figure~\ref{fig:main}, the \texttt{complete\_task} rules sit in block $C_2$,
which gets the most negative estimate and is dropped, although its lessons are tagged helpful. The budget is a cap, not a target: a playbook with one helpful block is curated to that block alone.
The rule is optimal when block effects add, $V(T) = V(\emptyset) + \sum_{j \in T} \beta_j$: then $\theta_j = \beta_j$, and without
a budget the positive blocks maximize $V$ over sets of blocks. Within a budget, largest first is the greedy knapsack solution, exact for blocks of equal size; greedy would rank unequal blocks by effect per token, but dividing a noisy estimate by a small size inflates its noise, so we rank by effect. Blocks that interact can be worth more or less together
than their estimates say, and one growth run may hold nothing of value; the last step checks both, scoring whole playbooks and no playbook.

\subsection{Verifying across growth runs}
\label{sec:select}
\bcc{} grows $K$ playbooks $P_1, \dots, P_K$, with the training tasks in $K$ orders, and curates each into
$\tilde S_{(k)}$. It then scores $2K + 1$ candidates on fresh practice episodes (Figure~\ref{fig:main}, step~3), each with its own reason: the curated sets, best when block effects add; the
whole playbooks, in case blocks help only together; and no playbook, in case every run harms. It serves the
candidate with the highest practice score $\pi(S)$, its mean success on them:
\begin{equation}
  \textstyle \hat S \;=\; \arg\max_{S \in \{\tilde S_{(1)}, \dots, \tilde S_{(K)},\, P_1, \dots, P_K,\, \emptyset\}} \, \pi(S) .
  \label{eq:pick}
\end{equation}
The choice thus reuses no episode of the estimates and touches no evaluation task. Since no playbook ($\emptyset$) is itself a candidate, the choice is safe:

\begin{proposition}
\label{prop:verify}
Let each of the $C$ candidates be scored on $n$ practice episodes with independent outcomes in $[0, 1]$. With
probability at least $1 - \delta$, the expected practice success of $\hat S$ is at most $2\sqrt{\log(2C/\delta)/(2n)}$ below the best candidate's, and hence below no playbook's.
\end{proposition}
The bound is on true success, which the winning noisy score overstates, and carries over to evaluation tasks as far
as they resemble the practice tasks (Section~\ref{sec:selection}). The margin shrinks as $1/\sqrt{n}$ but is loose at our sample sizes, about a
quarter of the scale (Appendix~\ref{app:verify}); in practice the drop is far smaller: on evaluation tasks, \bcc{} never falls more than \numKDropRbcThree{} points below no playbook on ACE (Table~\ref{tab:main}). RSEA's acceptance rule \citep{rsea2026}, which we call
\mBestFull, grows the same playbooks and serves the best whole one or no playbook by the same $\pi$: it does everything \bcc{} does except
curate, so their gap is what curating adds beyond choosing among runs.

\begin{table*}[t]
\centering\small\renewcommand{\arraystretch}{1.0}
\fittable{20}{%
\begin{tabular}{llrrrrrrrr}
\toprule
\multirow{2}{*}{\textbf{Model}} & \multirow{2}{*}{\textbf{Benchmark}} & \multicolumn{2}{c}{\textbf{Reference}} & \multicolumn{3}{c}{\textbf{Judging}} & \multicolumn{1}{c}{\textbf{Choosing}} & \multicolumn{2}{c}{\textbf{Measuring (ours)}} \\
\cmidrule(lr){3-4}\cmidrule(lr){5-7}\cmidrule(lr){8-8}\cmidrule(lr){9-10}
 & & \mNone & \mFull & \mOwn & \mTopk & \mRandom & \mBestFull & \mCut & \cellcolor{black!10}\bcc \\
\midrule
\arrayrulecolor{white}\specialrule{1pt}{0pt}{0pt}\arrayrulecolor{black}\multicolumn{10}{c}{\cellcolor{black!5}\textsc{ACE}} \\
Qwen 27B & AppWorld, pass/fail & \textbf{78.4} & 19.2 & 16.0 & 35.2 & 37.8 & \textbf{78.4} & 76.3 & \cellcolor{black!10}77.6 \\
 & AppWorld, per-test & 78.4 & 57.1 & 58.9 & 60.3 & 67.9 & 82.5 & 81.8 & \cellcolor{black!10}\textbf{84.3} \\
 & WebShop & 44.4 & 60.3 & 58.4 & 45.3 & 48.3 & 63.0 & 60.3 & \cellcolor{black!10}\textbf{63.1} \\
 & ALFWorld & 74.9 & 89.7 & 89.9 & 85.3 & 80.5 & \textbf{97.5} & 85.9 & \cellcolor{black!10}\textbf{97.5} \\
Qwen 9B & AppWorld, pass/fail & 19.8 & 21.7 & 20.7 & 24.5 & 21.5 & \textbf{40.1} & 25.7 & \cellcolor{black!10}\textbf{40.1} \\
 & AppWorld, per-test & 19.8 & 22.4 & 20.8 & 19.8 & 21.9 & \textbf{40.3} & 27.4 & \cellcolor{black!10}\textbf{40.3} \\
 & WebShop & 43.8 & 32.4 & 29.5 & 28.1 & 32.3 & 43.8 & 42.4 & \cellcolor{black!10}\textbf{49.6} \\
Gemma 31B & AppWorld, pass/fail & 70.6 & 48.8 & 52.9 & 49.1 & 51.4 & 65.3 & \textbf{70.8} & \cellcolor{black!10}69.8 \\
 & AppWorld, per-test & 69.9 & 70.5 & \textbf{71.9} & 70.6 & 70.0 & 70.1 & 71.2 & \cellcolor{black!10}70.1 \\
 & WebShop & 47.7 & 56.7 & 55.8 & 48.1 & 47.4 & 57.0 & 57.9 & \cellcolor{black!10}\textbf{58.1} \\
 & ALFWorld & 50.6 & 84.3 & 85.1 & 83.9 & 81.2 & \textbf{88.8} & 81.2 & \cellcolor{black!10}\textbf{88.8} \\
\midrule
\multicolumn{2}{l}{\textit{Average}} & 54.4 & 51.2 & 50.9 & 50.0 & 50.9 & 66.1 & 61.9 & \cellcolor{black!10}\textbf{67.2} \\
\multicolumn{2}{l}{\textit{Tokens served (\% of whole)}} & 0 & 100 & $\le$40 & $\le$40 & $\le$40 & 78 & 36 & \cellcolor{black!10}60 \\
\multicolumn{2}{l}{\textit{Largest drop below \mNone}} & -- & 69.6 & 72.6 & 60.3 & 60.6 & 5.4 & 12.2 & \cellcolor{black!10}\textbf{0.8} \\
\addlinespace[3pt]
\midrule
\arrayrulecolor{white}\specialrule{1pt}{0pt}{0pt}\arrayrulecolor{black}\multicolumn{10}{c}{\cellcolor{black!5}\textsc{Dynamic Cheatsheet}} \\
Qwen 27B & AppWorld, pass/fail & 78.4 & 83.1 & 80.3 & 82.6 & 82.2 & \textbf{83.7} & 81.0 & \cellcolor{black!10}80.4 \\
 & WebShop & 44.4 & 64.3 & 54.2 & 54.3 & 54.1 & \textbf{64.7} & 56.6 & \cellcolor{black!10}\textbf{64.7} \\
 & ALFWorld & 74.9 & 86.6 & 86.1 & 83.7 & 76.5 & \textbf{96.3} & 83.3 & \cellcolor{black!10}\textbf{96.3} \\
Qwen 9B & AppWorld, pass/fail & 19.8 & 16.0 & 17.1 & 17.9 & 18.6 & 15.3 & \textbf{20.1} & \cellcolor{black!10}15.3 \\
 & WebShop & 43.8 & 45.1 & 46.6 & 41.1 & 44.5 & \textbf{57.0} & 51.6 & \cellcolor{black!10}\textbf{57.0} \\
Gemma 31B & AppWorld, pass/fail & \textbf{70.6} & 64.6 & 65.7 & 65.7 & 66.5 & 63.5 & 64.7 & \cellcolor{black!10}63.5 \\
 & WebShop & 47.7 & 56.7 & 53.6 & 52.4 & 49.9 & \textbf{61.5} & 57.5 & \cellcolor{black!10}\textbf{61.5} \\
 & ALFWorld & 50.6 & 74.3 & 57.9 & 73.7 & 71.4 & \textbf{93.3} & 74.1 & \cellcolor{black!10}\textbf{93.3} \\
\midrule
\multicolumn{2}{l}{\textit{Average}} & 53.8 & 61.3 & 57.7 & 58.9 & 58.0 & \textbf{66.9} & 61.1 & \cellcolor{black!10}66.5 \\
\multicolumn{2}{l}{\textit{Tokens served (\% of whole)}} & 0 & 100 & $\le$40 & $\le$40 & $\le$40 & 100 & 35 & \cellcolor{black!10}93 \\
\multicolumn{2}{l}{\textit{Largest drop below \mNone}} & -- & 14.0 & 8.2 & 17.0 & 9.5 & \textbf{7.1} & 8.7 & \cellcolor{black!10}\textbf{7.1} \\
\midrule
\multicolumn{2}{l}{\textit{Both systems}} & 54.1 & 55.5 & 53.8 & 53.8 & 53.9 & 66.4 & 61.6 & \cellcolor{black!10}\textbf{66.9} \\
\bottomrule
\end{tabular}}
\caption{Task success (0--100) per setting, averaged over three evaluation seeds and the three growth runs; \mBestFull{} and \bcc{} instead choose one playbook among the runs on practice tasks. Judging rules cut a playbook to 40\% of its tokens (\mOwn: by ACE's tags or Dynamic Cheatsheet's usage count). Tokens served: in percent of the whole playbook, averaged over the runs. Largest drop: the most a method falls below \mNone{} in any setting and run. Best per row in bold.}
\label{tab:main}
\end{table*}

\section{Experiments}
\label{sec:experiments}

\noindent We fixed every constant of \bcc{} on four ACE settings, using none of their evaluation tasks, and kept it
unchanged on the other fifteen. \bcc{} grows \mbox{$K = 3$} playbooks per setting, the runs of Figure~\ref{fig:runs},
and curates each to a budget $B$ of 40\% of its tokens; 20\% and 60\% scored lower
(Appendix~\ref{app:sensitivity}). Every score averages three evaluation seeds over tasks, and over
the three growth runs where a method serves one run, so each score rests on hundreds to thousands of episodes.

\tpar{Baselines.}Every method serves from the same playbooks, on the same tasks and seeds, and differs only in
how it decides what to serve. Besides the reference points \mNone{} and \mFull, three baselines \emph{judge} and cut
the playbook to the same budget: by the system's own judgement (\mOwn: ACE's tags or Dynamic Cheatsheet's usage
count), by similarity to the task as in ReasoningBank (\mTopk; \citealp{ouyang2025reasoningbank}), or at random
(\mRandom). The strongest baseline \emph{chooses}: \mBestFull, the acceptance rule of RSEA \citep{rsea2026}, serves the best whole playbook or no playbook on practice tasks. Ours \emph{measure}: \mCut{} curates one growth run, and
\bcc{} adds the choice. Baselines run at their defaults (Appendix~\ref{app:baselines}).

\subsection{Measuring before serving lifts both memory systems on every model}
\label{sec:main}
Over all \numSysAllSettings{} settings, \bcc{} averages
\numSysAllRbc{} task success against \numSysAllFull{} with the whole playbook and \numSysAllNone{} with no playbook,
the best of all methods (Table~\ref{tab:main}). The gain is broad, not carried by a few settings: \bcc{} is above the
whole playbook in \numSysAboveFullAll{} of the \numSysAllSettings{} and above both reference points for every model
on both systems. Untuned on Dynamic Cheatsheet, \bcc{} lifts success there to \numSysDcRbc{} from \numSysDcFull{} (whole playbook)
and \numSysDcNone{} (no playbook).

\tpar{Little harm remains on ACE.}The whole playbook falls up to \numKDropFull{} points below no playbook, and every
baseline, even \mBestFull, at least \numMainBaselineDropMin{} below it somewhere; \bcc{} at most
\numKDropRbcThree, within evaluation noise. On Dynamic Cheatsheet, it shares
\mBestFull's two AppWorld drops (Table~\ref{tab:main}).

\subsection{Measuring beats judging}
\label{sec:strong}
\begin{figure}[!h]
\centering
\includegraphics[width=\columnwidth]{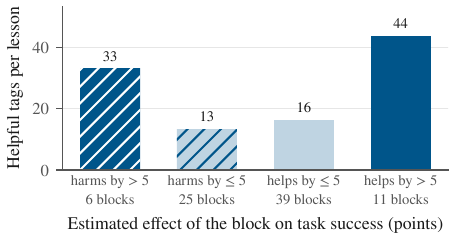}
\caption{The agent's tags against the measured effect of a block: median helpful tags per lesson for the \numTagsBlocks{} blocks of the ten ACE settings with tagged lessons.}
\label{fig:tageffect}
\end{figure}

Cutting a playbook by judgement keeps its harm. At the same budget, the system's own judgement, similarity to the task
and random choice score no better than the whole playbook (at most \numMainRuleAvgMax{} on ACE against
\numKAvgFull, \numMainRuleAvgMaxDc{} on Dynamic Cheatsheet against \numSysDcFull), while measuring one growth run
reaches \numKAvgCut{} on ACE (\mCut; Table~\ref{tab:main}); each judge falls at least \numMainRuleDropMin{} points
below no playbook in some ACE setting, the agent's own tags furthest (\numMainOwnDropAce). Judgement fails because it does not track what a block does
(Figure~\ref{fig:tageffect}): over \numTagsBlocks{} blocks, ACE's tags and the measured effects have a rank correlation
of only \numTagsRho, and the blocks that harm most carry more helpful tags than all but the most helpful ones. Judges given an advantage (similarity tuned on evaluation tasks, an LLM curator told each outcome) still stay
below \mCut, and, from the same episodes, DG-Mem's subset-Shapley estimates curate worse than \bcc's
(Appendix~\ref{app:strong}).

\subsection{Curating adds to choosing the best whole playbook, with fewer tokens}
\label{sec:systems}
Choosing among growth runs, as \mBestFull{} does, already avoids most of the harm; curating adds by keeping the helpful blocks of a harmful playbook. On ACE, \bcc{} scores \numKAvgRbcThree{} against
\numKAvgBestFullThree{} while serving \numMainTokRbcAce\% of the whole playbook's tokens against \numMainTokBfAce\%,
leads in \numSelAhead{} of the \numSelPicksDiffer{} settings where their choices differ, and by \numSelNonDevGain{}
on the seven settings not used to fix it. It gains most where all whole playbooks harm: \numMainRbcNineWs{}
against \numMainBestFullNineWs{} on 9B WebShop and \numMainRbcGemmaAwPf{} against \numMainBestFullGemmaAwPf{} on Gemma
AppWorld (pass/fail).

\begin{figure}[!h]
\centering
\includegraphics[width=\columnwidth]{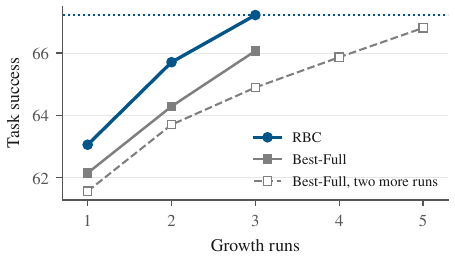}
\caption{Mean task success over the eleven ACE settings, averaged over every choice of growth runs from the three of Table~\ref{tab:main} (solid) or from five (dashed; the two added runs are weaker). Dotted: \bcc{} with three.}
\label{fig:frontier}
\end{figure}

\tpar{More runs do not replace curating.}\label{sec:cost}\label{sec:ablate}\bcc{} is ahead of \mBestFull{} at every
number of runs (Figure~\ref{fig:frontier}). Choosing needs two more, weaker runs to catch up (\numAkBestFullFive{}
against \numKAvgRbcThree) and serves \numAkServePct\% more tokens; at matched growth and practice cost, \bcc{} on two runs
and \mBestFull{} on three are level (\numKAvgRbcTwo{} against \numKAvgBestFullThree), and \mBestFull{} serves
\numAkServePctMatched\% more.

\tpar{Each part adds.}Curating one run alone scores \numKAvgCut{} on ACE and choosing alone
\numKAvgBestFullOne; both on one run score \numKAvgRbcOne, and over three runs \numKAvgRbcThree{} (Table~\ref{tab:app-parts}), with the
same order on Dynamic Cheatsheet.

\tpar{It curates only where curating pays.}Dynamic Cheatsheet rewrites its playbook as a whole, so its lessons are
less separable, and a curated cheatsheet scores level with the whole one (\numSysDcCut{} against \numSysDcFull).
\bcc{} detects this on practice tasks and serves a whole playbook in \numSysDcPicksWhole{} of
\numSysDcSettings{} settings, the same one as \mBestFull, so the two are level (\numSysDcRbc{} against \numSysDcBestfull): it falls back on
choosing.

\subsection{Curating pays most when playbooks grow long}
\label{sec:long}
\begin{table}[!h]
\centering\small
\fittable{10}{%
\begin{tabular}{@{}lrrrrr@{}}
\toprule
\hbm{Setting} & \hg{4}{Task success} & \hdmR{Tokens} \\
\cmidrule(lr){2-5}
 & \hs{\mNone} & \hs{\mFull} & \hs{\mBestFull} & \hs{\cellcolor{black!10}\bcc} & \\
\midrule
27B AW p/f & 78.4 & 64.0 & \textbf{86.3} & \cellcolor{black!10}\textbf{86.3} & 108 \\
27B AW p-t & 78.4 & 46.6 & 78.4 & \cellcolor{black!10}\textbf{84.9} & 31 \\
27B WS & 44.4 & 61.1 & 65.0 & \cellcolor{black!10}\textbf{65.1} & 44 \\
27B ALF & 74.9 & 83.6 & \textbf{98.5} & \cellcolor{black!10}\textbf{98.5} & 98 \\
\midrule
9B AW p/f & 19.8 & 28.2 & 31.0 & \cellcolor{black!10}\textbf{34.7} & 36 \\
9B AW p-t & 19.8 & 17.7 & 21.2 & \cellcolor{black!10}\textbf{32.3} & 34 \\
\midrule
Gemma AW p/f & 70.6 & 67.4 & 67.7 & \cellcolor{black!10}\textbf{73.6} & 43 \\
Gemma AW p-t & 69.9 & 70.3 & 70.8 & \cellcolor{black!10}\textbf{71.6} & 41 \\
Gemma ALF & 50.6 & 87.8 & \textbf{90.5} & \cellcolor{black!10}\textbf{90.5} & 89 \\
\bottomrule
\end{tabular}}
\caption{Long growth on ACE: up to ten passes, three new growth runs per setting; whole playbooks end 1.0--2.3 times their standard size. Tokens: what \bcc{} serves, in \% of the average whole playbook. AW: AppWorld (p/f, p-t feedback), WS: WebShop, ALF: ALFWorld.}
\label{tab:long}
\end{table}

ACE never removes a lesson, so playbooks keep growing. We grew three runs for up to ten passes on nine settings,
leaving out WebShop for 9B and Gemma, since the 27B WebShop playbook barely grew (Table~\ref{tab:long}). On AppWorld
with per-test feedback, whole playbooks then stop helping (\numLgTwentysevenAwPtFull{} against
\numLgTwentysevenAwPtNone{} with no playbook for 27B, \numLgNineAwPtFull{} against \numLgNineAwPtNone{} for 9B). Choosing can then only fall back on no playbook or a weak run,
while \bcc{} still finds the helpful blocks: \numLgTwentysevenAwPtRbc{} against \numLgTwentysevenAwPtBf{} for
\mBestFull{} on 27B and \numLgNineAwPtRbc{} against \numLgNineAwPtBf{} on 9B. Where a whole playbook is best, \bcc{}
serves the same one as \mBestFull; in the other \numLgCutN{} settings it serves a curated set, at most \numLgCutTokMax\% of the tokens,
and scores \numLgCutGain{} higher on average. Its lead of \numLgMeanGain{} over \mBestFull{} thus comes
entirely from curating, not from choosing.

\subsection{Practice tasks should come from the tasks the agent will face}
\label{sec:selection}
Lessons are general rules, such as how to end a task, so choosing on practice tasks carries over to evaluation
tasks when the two are alike: practice
success ranks the seven candidates of a setting much as evaluation success does (median Spearman correlation
\numKRho), and \bcc{} picks the best candidate on evaluation in \numKPickBest{} of the \numKPickN{} ACE settings and
one within half a point of it in \numKPickWithinHalf. Where it misses, the tasks differ: AppWorld's practice tasks come
from another split. This largely explains the two Dynamic Cheatsheet AppWorld settings that stay below no playbook
(9B and Gemma, pass/fail). Choosing on held-out tasks of the evaluation split
instead raises \bcc{} by a further \numCrossDcGain{} points on Dynamic Cheatsheet and \numCrossAceGain{} points on ACE
(Appendix~\ref{app:crossfit}).

\section{Conclusion}
\label{sec:conclusion}

In this work, we showed that self-written playbooks are unstable: one procedure can grow a helpful or a harmful
playbook, and the agent's own judgement cannot tell which. We proposed \bcc{}, which measures instead:
it serves blocks of lessons at random on practice tasks to see what each does, and checks the curated set against
whole playbooks and no playbook before serving. Task success rises on average from \numSysAllFull{} with the whole playbook to
\numSysAllRbc, and curating adds to choosing with fewer tokens, above all under long growth. An agent's memory
should be measured, not judged.

\section*{Limitations}
\label{sec:limitations}

Our evidence covers two memory systems, three benchmarks and three open-weight models, with playbooks we grew;
other memory systems, domains and larger or proprietary models remain to be tested. Within this scope, \bcc{} values a block by its
average gain over the other blocks served with it, which makes curation exact when block effects add; when lessons
are hard to separate, as in a memory rewritten as a whole, it falls back on the best whole playbook, which still lifts
success but saves no context. \bcc{} also trades extra episodes for reliability: it grows several playbooks, runs
practice rounds and needs practice tasks with an observable outcome. Since its guarantee holds on these practice
tasks, it works best when they resemble the tasks the agent will face. We fixed its constants on four ACE settings
and kept them unchanged elsewhere. Finally, we study playbooks fixed once grown; extending \bcc{} to memory that grows
while it is served is a natural next step.

\bibliography{refs}

\appendix
\makeatletter\g@addto@macro\normalsize{\setlength\abovedisplayskip{4pt plus 1pt minus 1pt}\setlength\belowdisplayskip{4pt plus 1pt minus 1pt}\setlength\abovedisplayshortskip{1pt plus 1pt}\setlength\belowdisplayshortskip{3pt plus 1pt}}\makeatother\normalsize
\section{Experimental setup}
\label{app:ace}
\begin{table*}[t]
\centering\small\setlength{\tabcolsep}{6pt}
\fittable{6}{%
\begin{tabular}{@{}llll@{}}
\toprule
 & \hb{AppWorld} & \hb{WebShop} & \hb{ALFWorld} \\
\midrule
Models & 27B, 9B, Gemma & 27B, 9B, Gemma & 27B, Gemma \\
Memory systems & ACE (both feedback types), DC & ACE, DC & ACE, DC \\
Score (0--100) & task goal completion & purchase reward & success \\
Actor & ReAct, 40 steps & ReAct, 15 actions & ReAct, 50 steps \\
Feedback during growth & pass/fail, or per-test & reward and its components & success \\
Growth tasks & 90 train tasks, batch 16 & 500 train instructions, batch 32 & 500 train games, batch 32 \\
Practice tasks & dev split, 57 tasks & 300 instructions & \texttt{eval\_seen}, 140 games \\
Practice round & 684 episodes & 600 episodes & 600 episodes \\
Evaluation tasks & \texttt{test\_normal}, 168 $\times$ 3 seeds & test, 500 $\times$ 3 seeds & \texttt{valid\_unseen}, 134 $\times$ 3 seeds \\
\bottomrule
\end{tabular}}
\caption{Settings and protocol: eleven for ACE (AppWorld counts once per feedback type) and eight for Dynamic
Cheatsheet (DC; AppWorld with pass/fail feedback only). Practice and evaluation tasks never overlap; WebShop's
practice instructions exclude its 500 test instructions. Actors sample at temperature 0.7, all other roles at 0.}
\label{tab:cells}
\end{table*}

\subsection{ACE as we run it}
\label{app:ace-recipe}
\tpar{A playbook.}Each line of an ACE playbook is one lesson with an id and the agent's tags, for example:
\begin{quote}\small
\texttt{[sai-00001] helpful=166 harmful=0} :: Always extract dynamic identifiers (e.g., \texttt{artist\_id}, \texttt{track\_id}) from the output of previous API calls and use them in subsequent requests. Never assume or hardcode IDs unless they are explicitly static and known in advance.
\end{quote}
The actor sees the lesson without the tags.

\tpar{Growth.}We run ACE's batched mode \citep{zhang2025ace}. On each batch of training tasks, the actor plays every
task with the current playbook; the Reflector reads each trajectory with its feedback and tags the lessons that bore
on it as helpful, harmful or neutral; the Curator writes new lessons, which are appended; and near-duplicate lessons
are merged. No step removes a lesson, and the whole playbook is served on every call (\mFull). Growth runs differ in
the order of the training tasks and the actor's sampling; a run stops once a pass over the training tasks grows the
playbook by less than 10\%, after at most five passes.

\tpar{Feedback.}On AppWorld, the Reflector sees either whether the task passed (pass/fail) or the result of each unit
test (per-test); the two grow different playbooks and count as different settings. On WebShop it sees the reward,
on ALFWorld success.

\subsection{Dynamic Cheatsheet as we run it}
\label{app:dc}
We run Dynamic Cheatsheet's cumulative mode \citep{suzgun2025dynamic} with the actors and training tasks of our ACE
runs. After each batch of training tasks, one model call reads the current cheatsheet and the batch's trajectories,
judges by itself, without the reward, which attempts succeeded, and writes a new cheatsheet that replaces the old one,
with a count of each entry's successful uses. The cheatsheet is served whole; \bcc{} treats its entries as lessons and is otherwise unchanged.

\subsection{Settings, tasks and models}
\label{app:settings}
Table~\ref{tab:cells} lists the settings and their tasks. Every method chooses on practice tasks only, and practice
tasks are never scored. Scores are on 0--100, averaged over seeds per task and then over tasks. \bcc{} was developed
on four ACE settings, scored on tasks apart from their evaluation tasks: 27B on AppWorld with pass/fail feedback and on WebShop, and 9B on AppWorld with both feedback
types.

\tpar{Models.}Qwen3.8-27B (27B), Qwen3.5-9B (9B) and Gemma-4-31B (Gemma) play every role in their own settings, with
thinking off. Gemma is evaluated on an FP8 checkpoint \citep{micikevicius2022fp8}, and its growth runs also differ in
precision.

\subsection{Baselines}
\label{app:baselines}
\begin{table*}[t]
\centering\small
\begin{tabular*}{\textwidth}{@{\extracolsep{\fill}}lrrrrrrrrrr@{}}
\toprule
\hbm{Block} & \hdm{Effect} & \hdm{Lessons} & \hdm{Tokens} & \hg{2}{Tags} & \hdm{Top half} & \hg{2}{Success} & \multicolumn{2}{c@{}}{\hb{Answer / buy}} \\
\cmidrule(lr){5-6}\cmidrule(lr){8-9}\cmidrule(l){10-11}
 & & & & \hs{Helpful} & \hs{Harmful} & & \hs{In} & \hs{Out} & \hs{In} & \hsR{Out} \\
\midrule
\multicolumn{11}{@{}l}{\textit{27B, AppWorld, pass/fail feedback}} \\
\quad b00 & $-$31.8 & 14 & 1{,}237 & 457 & 10 & 64.3 & 25.0 & 53.9 & 69.0 & 37.5 \\
\quad b04 & $-$12.9 & 21 & 2{,}276 & 701 & 141 & 52.4 & 33.3 & 43.2 & 60.3 & 48.8 \\
\addlinespace[2pt]
\multicolumn{11}{@{}l@{}}{\hspace{1em}\rlap{\parbox[t]{\dimexpr\textwidth-1em\relax}{\footnotesize Rule in b00, tagged \numAnatTwentysevenAwPfBZeroRuleHelp{} times helpful and \numAnatTwentysevenAwPfBZeroRuleHarm{} times harmful: ``When calling \texttt{complete\_task}, always pass a concrete, non-empty string literal as the answer argument. Keep the string concise and factual (e.g., `Removed 5 songs' or the specific ID/name) to avoid parsing failures, length limit issues, or validation errors caused by verbose explanations or variable references.''\strut}}} \\
\midrule
\multicolumn{11}{@{}l}{\textit{9B, WebShop}} \\
\quad b02 & $-$12.8 & 10 & 1{,}027 & 133 & 586 & 20.0 & 18.3 & 25.6 & 27.0 & 38.5 \\
\quad b06 & $-$6.9 & 6 & 769 & 116 & 25 & 33.3 & 20.1 & 23.5 & 27.0 & 37.7 \\
\addlinespace[2pt]
\multicolumn{11}{@{}l@{}}{\hspace{1em}\rlap{\parbox[t]{\dimexpr\textwidth-1em\relax}{\footnotesize Rule in b02, tagged \numAnatNineWsBTwoRuleHelp{} times helpful and \numAnatNineWsBTwoRuleHarm{} times harmful: ``Do not assume qualitative attributes (e.g., `zipper pockets', `organic', `sugar free') or specific design options are present just because keywords appear in the title. \ldots''\strut}}} \\
\midrule
\multicolumn{11}{@{}l}{\textit{9B, AppWorld, pass/fail feedback}} \\
\quad b01 & $-$4.3 & 17 & 1{,}924 & 102 & 50 & 41.2 & 13.6 & 20.4 & 78.7 & 66.8 \\
\quad b04 & $-$5.2 & 15 & 1{,}668 & 271 & 920 & 66.7 & 13.0 & 21.6 & 75.1 & 69.5 \\
\addlinespace[2pt]
\multicolumn{11}{@{}l@{}}{\hspace{1em}\rlap{\parbox[t]{\dimexpr\textwidth-1em\relax}{\footnotesize Rule in b04, tagged \numAnatNineAwPfBFourRuleHelp{} times helpful and \numAnatNineAwPfBFourRuleHarm{} times harmful: ``For the \texttt{complete\_task} API, strictly pass only primitive types: a raw integer count, a boolean, or a minimal status string (e.g., `Done', `Task completed'). \ldots''\strut}}} \\
\midrule
\multicolumn{11}{@{}l}{\textit{Gemma, AppWorld, pass/fail feedback}} \\
\quad b00 & $-$6.5 & 21 & 1{,}350 & 967 & 2 & 28.6 & 25.0 & 33.3 & 62.2 & 58.3 \\
\quad b02 & $-$16.6 & 21 & 1{,}435 & 686 & 0 & 66.7 & 22.7 & 33.9 & 69.3 & 53.2 \\
\quad b03 & $-$24.0 & 19 & 1{,}245 & 710 & 0 & 63.2 & 16.1 & 41.7 & 76.2 & 44.6 \\
\addlinespace[2pt]
\multicolumn{11}{@{}l@{}}{\hspace{1em}\rlap{\parbox[t]{\dimexpr\textwidth-1em\relax}{\footnotesize Rule in b03, tagged \numAnatGemmaAwPfBThreeRuleHelp{} times helpful and \numAnatGemmaAwPfBThreeRuleHarm{} times harmful: ``Treat the final completion API call (e.g., complete\_task) as the primary submission mechanism; ensure the `answer' or `summary' parameter is populated with the actual result of the request to avoid it being interpreted as a failure to provide output.''\strut}}} \\
\bottomrule
\end{tabular*}
\caption{The harmful blocks of the four harmful settings: estimated effect negative at a 5\% false-discovery rate over the eight blocks (first practice round). Effect: points of success with the block included minus left out. Tags: summed over the block's lessons. Top half: \% of its lessons in the playbook's top half by helpful tags (chance 50). In / Out: unadjusted rates (\%) on the episodes that included or left out the block; Answer / buy: answering a task that expects no answer (AppWorld) or buying (WebShop). Long rules are quoted up to their first sentence.}
\label{tab:anatomy}
\end{table*}

Every budgeted method draws from the same lessons, fills the same budget $B$, and is scored on the same tasks and
seeds. Baselines run at their defaults and are never tuned, except the two marked $^{*}$, which are tuned on the
evaluation tasks on purpose.

\tpar{\mRandom.}Random lessons until $B$ is full; mean of two orders.

\tpar{\mOwn.}On ACE (\mCounters), lessons ranked by their share of helpful tags; on Dynamic Cheatsheet, by usage
count. This is the cut the system's own judgement implies.

\tpar{\mTopk.}The lessons most similar to the task text, as in ReasoningBank \citep{ouyang2025reasoningbank}.

\tpar{\mCountersT{} and \mTopkT.}Per setting, the best of eight tag rules or of five values of $k$, chosen on the
evaluation tasks, which can only favour them.

\tpar{\mDCs{} and \mDCo.}Dynamic Cheatsheet's curator, run over the fixed lessons on practice episodes to choose the
lessons to keep within $B$. \mDCs{} judges success by itself; \mDCo{} is told the task outcome, the signal \bcc{}
uses.

\tpar{\mBestFull.}RSEA's acceptance rule \citep{rsea2026} across $K$ runs: of no playbook and the $K$ whole
playbooks, the one with the highest practice success, on the same runs and practice episodes as \bcc{}
(Algorithm~\ref{alg:bcck}).

\tpar{DG-Mem's estimates.}DG-Mem's subset-Shapley estimate \citep{dgmem2026}, computed per block from the same
practice episodes as \bcc's regression on the four 27B settings and served by \bcc's rule; only the estimator
differs.

\section{Analysis details}
\label{app:analysis}

\subsection{The harmful blocks}
\label{app:anatomy}
Table~\ref{tab:anatomy} lists every harmful block of the four harmful settings with one of its rules. The harm is not
scattered: in each setting it comes from one family of rules that sound useful.

\tpar{AppWorld: one family of rules.}On all three AppWorld pass/fail settings the harmful blocks tell the agent to
pass an answer to \texttt{complete\_task}, the call that ends the task, even when the task expects none. With such a
block, the agent answers more of these tasks and fails the check ``answers match'' more often. Under
pass/fail feedback the Reflector never sees the checker and writes rules for an imagined validator; under per-test
feedback it sees the failed tests and writes none.

\tpar{WebShop: a different family.}On 9B WebShop the harmful blocks make the agent verify a product again or
abandon it, and with block b02 included it buys less often.

\tpar{The tags.}On 27B and Gemma the harmful lessons carry mostly helpful tags and sit near chance in the tag
ranking (top-half share). On 9B, by contrast, the Reflector tags its harmful rules mostly as harmful. The rates are unadjusted means over practice episodes.

\subsection{Dynamic Cheatsheet playbooks across growth runs}
\label{app:dc-runs}
\begin{figure}[!h]
\centering
\includegraphics[width=\columnwidth]{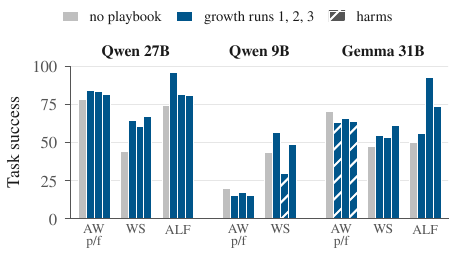}
\caption{Task success of the three growth runs of Dynamic Cheatsheet per setting, drawn as Figure~\ref{fig:runs}. Hatched: more than \numRunsHarmPts{} points below no playbook. AW p/f: AppWorld with pass/fail feedback; WS: WebShop; ALF: ALFWorld.}
\label{fig:app-runs-dc}
\end{figure}

A playbook that is rewritten as a whole depends on its growth run just as one grown lesson by lesson
(Figure~\ref{fig:app-runs-dc}, which repeats Figure~\ref{fig:runs} for Dynamic Cheatsheet). Its playbooks help on
average (\numSysDcFull{} against \numSysDcNone{} with no playbook), but the runs of one setting lie up to
\numSysDcSpread{} points apart (\numSysDcSpreadWhere), and on 9B WebShop one run helps while another harms. The
instability of Section~\ref{sec:analysis} is thus a property of self-written memory, not of one system.

\section{Pseudocode}
\label{app:pseudocode}

Algorithms~\ref{alg:bcc} and~\ref{alg:bcck} give curation of one growth run and the full method of
Section~\ref{sec:method} as our code runs them. A lesson $b$ has a token cost $c(b)$ and a usage rate $u(b)$, how often the Reflector tagged it during growth,
which only orders lessons within a block; $c(S) = \sum_{b \in S} c(b)$. The constants are the same in every setting:
$m = 8$, $p = \tfrac12$, a budget share $\rho = 0.4$, $K = 3$, and a grid $\Lambda$ of ridge penalties that
includes $0$.

\begin{algorithm}[!ht]
\caption{Curating one growth run (\mCut)}
\label{alg:bcc}
\small
\begin{algorithmic}[1]
\Require playbook $P$, budget $B$, practice tasks $x_1, \dots, x_n$
\Ensure curated set $\tilde S \subseteq P$
\State $C_1, \dots, C_m \gets$ Ward clusters of lesson embeddings
\For{$i = 1, \dots, n$} \Comment{practice round}
  \State draw $z_{ij} \sim \mathrm{Bernoulli}(p)$ independently, $j \in [m]$
  \State $y_i \gets$ success on $x_i$ when served $\bigcup_{j : z_{ij} = 1} C_j$
\EndFor
\For{$\lambda \in \Lambda$} \Comment{effects}
  \State $\hat\theta(\lambda) \gets$ minimizer of Eq.~\eqref{eq:ridge} with penalty $\lambda$
\EndFor
\State $\hat\theta \gets \hat\theta(\lambda^{\star})$, $\lambda^{\star}$ best in leave-one-task-out error
\State $L \gets \bigcup_{j : \hat\theta_j > 0} C_j$, by $\hat\theta_j$ then $u(b)$, descending
\State $\tilde S \gets \emptyset$ \Comment{$L = \emptyset$: no playbook}
\ForAll{$b \in L$ in order}
  \If{$c(\tilde S) + c(b) \le B$} $\tilde S \gets \tilde S \cup \{b\}$ \EndIf
\EndFor
\State \Return $\tilde S$, in playbook order
\end{algorithmic}
\end{algorithm}

\begin{algorithm}[!ht]
\caption{\bcc{} and the reference \mBestFull}
\label{alg:bcck}
\small
\begin{algorithmic}[1]
\Require training tasks, practice tasks $\Dpr$
\Ensure served sets of \bcc{} and \mBestFull
\For{$k = 1, \dots, K$}
  \State $P_k \gets$ growth, training tasks in order $k$
  \State $\tilde S_{(k)} \gets \mathrm{Curate}(P_k,\ \rho\, c(P_k))$ \Comment{Algorithm~\ref{alg:bcc}}
\EndFor
\State $\pi(S) :=$ mean success on new episodes of $\Dpr$ served $S$
\State $\mathcal{C} \gets (\tilde S_{(1)}, \dots, \tilde S_{(K)}, P_1, \dots, P_K, \emptyset)$ \Comment{$2K + 1$ candidates}
\State $\hat S \gets \arg\max_{S \in \mathcal{C}} \pi(S)$ \Comment{ties: first in $\mathcal{C}$}
\State $S_{\mathrm{BF}} \gets \arg\max_{S \in \{\emptyset, P_1, \dots, P_K\}} \pi(S)$ \Comment{\mBestFull}
\State \Return $\hat S$, $S_{\mathrm{BF}}$
\end{algorithmic}
\end{algorithm}

Evaluation tasks enter neither algorithm.

\section{Proofs}
\label{app:proofs}
\subsection{Proposition~\ref{prop:estimate}: what the estimates measure}
\label{app:banzhaf}

One lemma does most of the work, and the three parts of the proposition follow from it: identification, precision
and the effect of the ridge penalty.

\tpar{Notation.}Practice tasks are drawn from $\mathcal D$, and $\mu(x, T)$ is the expected success on task $x$ when
the blocks in $T \subseteq [m]$ are served, so $V(T) = \mathbb{E}_{x \sim \mathcal D}\, \mu(x, T)$. Episodes are
i.i.d.: on episode $i$ the task is $x_i \sim \mathcal D$, the inclusions $z_{i1}, \dots, z_{im}$ are independent
$\mathrm{Bernoulli}(p)$ variables, independent of $x_i$, and $\mathbb{E}[\,y_i \mid x_i, z_i\,] = \mu(x_i, T_i)$ with
$T_i = \{ j : z_{ij} = 1 \}$. We write $\tilde y = y - \alpha(x)$ for success with the task mean
$\alpha(x) = \mathbb{E}[\,y \mid x\,]$ removed, and $\sigma^2$ for the largest $\mathrm{Var}(\tilde y \mid z_j = b)$
over blocks $j$ and $b \in \{0, 1\}$. For a block $j$, the other blocks included form a set
$T \subseteq [m] \setminus \{j\}$ with probability
\begin{equation*}
  w_p(T) = p^{|T|} (1-p)^{m-1-|T|} .
\end{equation*}

\begin{lemma}
\label{lem:contrast}
For every block $j$ and every $V$,
\begin{equation}
\begin{split}
  \Delta_j &:= \mathbb{E}[\,y \mid z_j{=}1\,] - \mathbb{E}[\,y \mid z_j{=}0\,] \\
  &\phantom{:}= \sum_{T} w_p(T) \big[ V(T \cup \{j\}) - V(T) \big] ,
\end{split}
  \label{eq:app-contrast}
\end{equation}
where $T$ runs over the subsets of $[m] \setminus \{j\}$.
\end{lemma}

\begin{appproof}
Tasks and inclusions are independent, so given $z_j = b$ the other blocks served form $T$ with probability
$w_p(T)$ and the task is still drawn from $\mathcal D$. Averaging over $T$,
\begin{align*}
  \mathbb{E}[\,y \mid z_j{=}1\,] &= \sum_{T} w_p(T)\, V(T \cup \{j\}) , \\
  \mathbb{E}[\,y \mid z_j{=}0\,] &= \sum_{T} w_p(T)\, V(T) ,
\end{align*}
and subtracting the two lines gives Eq.~\eqref{eq:app-contrast}.
\end{appproof}

\begin{appproof}[Identification]
At $p = \tfrac12$ every weight is $2^{-(m-1)}$, the uniform distribution over the subsets of the other blocks, so
by Eq.~\eqref{eq:estimand}
\begin{equation*}
  \Delta_j = \mathbb{E}_T \big[ V(T \cup \{j\}) - V(T) \big] = \theta_j ,
\end{equation*}
whatever $V$ is. The regression estimates the same quantity. By the Frisch--Waugh--Lovell theorem, and since the
inclusions are independent of the task and of one another, the coefficient on $z_j$ in the population fit with task
intercepts is $\mathrm{Cov}(z_j, y) / p(1-p)$. With $\eta_b = \mathbb{E}[\,y \mid z_j{=}b\,]$,
\begin{align*}
  \mathrm{Cov}(z_j, y) &= p\,\eta_1 - p \big( p\,\eta_1 + (1-p)\,\eta_0 \big) \\
  &= p(1-p)\, \Delta_j ,
\end{align*}
so the coefficient is $\Delta_j = \theta_j$. With finitely many tasks the sample fit converges to this fit, so
least squares ($\lambda = 0$ in Eq.~\eqref{eq:ridge}) is consistent for $\theta_j$.
\end{appproof}

\begin{appproof}[Precision]
Let $\hat\Delta_j$ be the mean of $\tilde y$ over the $n_1$ episodes that include block $j$ minus its mean over
the $n_0$ that leave it out. Given which episodes include $j$, the $\tilde y_i$ are independent and the task term
cancels, so $\hat\Delta_j$ is unbiased for $\Delta_j = \theta_j$, with
\begin{equation*}
  \mathrm{Var}(\hat\Delta_j) \le \sigma^2 \Big( \frac{1}{n_1} + \frac{1}{n_0} \Big) \approx \frac{\sigma^2}{n\, p(1-p)} ,
\end{equation*}
as $n_1 \approx pn$. Since $p(1-p) \le \tfrac14$, this is smallest at $p = \tfrac12$: $4\sigma^2 / n$ for every block,
from the same $n$ episodes. Leaving out one block at a time instead runs $m + 1$ arms (all blocks, and all but $j$
for each $j$) of $n / (m+1)$ episodes each, so each contrast has variance up to
\begin{equation*}
  \frac{2(m+1)\,\sigma^2}{n} = \frac{m+1}{2} \cdot \frac{4\sigma^2}{n} ,
\end{equation*}
the factor of the proposition \citep{box2005statistics}. The argument treats $\alpha$ as known; estimating it from
the same episodes does not change the first-order variance when $n$ grows with the tasks fixed.
\end{appproof}

\begin{table*}[t]
\centering\small\setlength{\tabcolsep}{6pt}
\fittable{22}{%
\begin{tabular}{lrrrrrrrrrr}
\toprule
\hbm{Setting} & \hdm{\mNone} & \hg{3}{\mFull, growth run} & \hg{3}{\mCut, growth run} & \hg{3}{Verified over three runs} \\
\cmidrule(lr){3-5}\cmidrule(lr){6-8}\cmidrule(l){9-11}
 & & \hs{1} & \hs{2} & \hs{3} & \hs{1} & \hs{2} & \hs{3} & \hs{\mBestFull} & \hs{Curated} & \hs{\cellcolor{black!10}\bcc} \\
\midrule
27B AppWorld p/f & 78.4 & 31.2 & 17.7 & 8.7 & 81.0 & 70.2 & 77.6 & \textbf{78.4} & 77.6 & \cellcolor{black!10}77.6 \\
9B WebShop & 43.8 & 24.3 & 34.5 & 38.3 & 45.9 & 31.6 & 49.6 & 43.8 & \textbf{49.6} & \cellcolor{black!10}\textbf{49.6} \\
9B AppWorld p/f & 19.8 & 13.5 & 40.1 & 11.5 & 22.2 & 40.3 & 14.7 & 40.1 & \textbf{40.3} & \cellcolor{black!10}40.1 \\
Gemma AppWorld p/f & 70.6 & 15.9 & 65.3 & 65.3 & 69.2 & 73.2 & 69.8 & 65.3 & \textbf{69.8} & \cellcolor{black!10}\textbf{69.8} \\
\midrule
27B WebShop & 44.4 & 59.0 & 58.9 & 63.0 & 63.0 & 54.7 & 63.1 & 63.0 & \textbf{63.1} & \cellcolor{black!10}\textbf{63.1} \\
27B ALFWorld & 74.9 & 97.5 & 93.8 & 77.9 & 95.3 & 84.6 & 77.9 & \textbf{97.5} & 95.3 & \cellcolor{black!10}\textbf{97.5} \\
27B AppWorld p-t & 78.4 & 82.5 & 58.9 & 30.0 & 84.3 & 80.4 & 80.8 & 82.5 & \textbf{84.3} & \cellcolor{black!10}\textbf{84.3} \\
9B AppWorld p-t & 19.8 & 40.3 & 17.3 & 9.5 & 44.0 & 21.4 & 16.7 & 40.3 & \textbf{44.0} & \cellcolor{black!10}40.3 \\
Gemma WebShop & 47.7 & 57.0 & 57.1 & 55.9 & 58.1 & 57.7 & 57.8 & 57.0 & \textbf{58.1} & \cellcolor{black!10}\textbf{58.1} \\
Gemma ALFWorld & 50.6 & 84.3 & 88.8 & 79.9 & 83.1 & 82.1 & 78.4 & \textbf{88.8} & 78.4 & \cellcolor{black!10}\textbf{88.8} \\
Gemma AppWorld p-t & 69.9 & 70.1 & 72.8 & 68.5 & 71.2 & 70.2 & 72.0 & 70.1 & \textbf{72.0} & \cellcolor{black!10}70.1 \\
\midrule
Mean & 54.4 & 52.3 & 55.0 & 46.2 & 65.2 & 60.6 & 59.8 & 66.1 & 66.6 & \cellcolor{black!10}\textbf{67.2} \\
\bottomrule
\end{tabular}
}
\caption{Every growth run (task success, 0--100); runs 2 and 3 regrow run 1 with the training tasks in other orders, and \mCut{} serves each run's curated set. The last three columns choose on practice tasks among no playbook and the whole playbooks (\mBestFull), the curated sets (Curated), or both (\bcc); best of the three in bold.}
\label{tab:app-runs}
\end{table*}

\begin{appproof}[Ridge keeps signs and order]
With the task means removed from $y$ and $z$ ($\tilde y$ and $\tilde Z$), the unpenalized intercepts drop out of
Eq.~\eqref{eq:ridge}, and the ridge estimate solves
\begin{equation*}
  (\tilde Z^{\top}\tilde Z + \lambda I)\, \hat\theta_\lambda = \tilde Z^{\top}\tilde y .
\end{equation*}
If $\tilde Z^{\top}\tilde Z = cI$ with $c > 0$, then $\hat\theta_\lambda = \tfrac{c}{c + \lambda}\, \hat\theta_0$:
every estimate is scaled by the same factor in $(0, 1]$, so signs and order are kept. Fair inclusion gives
$\mathbb{E}[\tilde Z^{\top}\tilde Z] \approx \tfrac n4 I$, so in a finite round only blocks with close estimates can swap order, and only estimates near zero can change sign.
\end{appproof}

\begin{corollary}
\label{cor:serve}
Suppose block effects add, that is, $V(T) = V(\emptyset) + \sum_{j \in T} \beta_j$ for every $T$. Then $\theta_j = \beta_j$ for every
$p$, and
\begin{enumerate}[label=(\alph*)]
  \item without a budget, $\{ j : \theta_j > 0 \}$ maximizes $V$;
  \item if every block costs $c_0$ tokens, the positive blocks with the largest $\theta_j$, at most
        $\lfloor B / c_0 \rfloor$ of them, maximize $V$ among the sets within the budget $B$.
\end{enumerate}
\end{corollary}

\begin{appproof}
Under additivity every bracket in Eq.~\eqref{eq:app-contrast} equals $\beta_j$, and the weights sum to one, so
$\theta_j = \beta_j$ and
\begin{equation*}
  V(T) - V(\emptyset) = \sum_{j \in T} \theta_j .
\end{equation*}
This sum is largest when $T$ holds every positive term, which proves (a), and, among sets of at most
$\lfloor B / c_0 \rfloor$ blocks, when $T$ holds the largest positive terms, which proves (b).
\end{appproof}

When effects do not add, $\theta_j$ still averages the interactions, but positive blocks can be worth less together
than their effects suggest (Appendix~\ref{app:sensitivity}).

\subsection{Proposition~\ref{prop:verify}: verification is safe}
\label{app:verify}
Let $S_1, \dots, S_C$ be the candidates, and let candidate $c$ be scored on $n$ practice episodes with outcomes
$y_{c1}, \dots, y_{cn} \in [0, 1]$, independent across episodes. Its practice score and expected practice success are
\begin{equation*}
  \pi_c = \frac1n \sum_{i=1}^{n} y_{ci} , \qquad V_{\mathrm{pr}}(S_c) = \mathbb{E}\, \pi_c .
\end{equation*}
By Hoeffding's inequality, for each candidate
\begin{equation*}
  \Pr\big( |\pi_c - V_{\mathrm{pr}}(S_c)| > \varepsilon \big) \le 2 \exp(-2 n \varepsilon^2) ,
\end{equation*}
and by the union bound over the $C$ candidates, all $C$ deviations are at most $\varepsilon$ with probability at least
$1 - 2C \exp(-2 n \varepsilon^2)$, which equals $1 - \delta$ for $\varepsilon = \sqrt{\log(2C/\delta)/(2n)}$. On
that event, let $\hat S$ be the candidate with the highest $\pi$ and $S_c$ any candidate. Then
\begin{equation*}
  V_{\mathrm{pr}}(\hat S) \ge \pi(\hat S) - \varepsilon \ge \pi(S_c) - \varepsilon \ge V_{\mathrm{pr}}(S_c) - 2\varepsilon ,
\end{equation*}
which holds for the best candidate and, since no playbook is one of them, for $S_c = \emptyset$.

The bound needs nothing about how the candidates were built, so it holds for any curation rule and any $K$; the
price of more candidates is the factor $\log C$. It is loose at our sample sizes: with $C = 7$ and the
\numCostScoreAppWorld{} practice episodes per candidate on AppWorld, $2\varepsilon$ at $\delta = 0.05$ is about a
quarter of the scale. In practice the choice lands much closer, within half a point of the best candidate on evaluation in
\numKPickWithinHalf{} of the eleven settings (Section~\ref{sec:selection}). The guarantee concerns the practice
tasks; it says nothing about evaluation tasks that differ from them.

\section{Further results}
\label{app:results}

\subsection{Per setting and per model}
\label{app:dc-results}\label{app:ace-all}
\begin{table}[!h]
\centering\small
\fittable{8}{%
\begin{tabular}{lrrrr}
\toprule
\hb{Model (settings)} & \hd{\mNone} & \hd{Whole} & \hd{\mBestFull} & \hd{\cellcolor{black!10}\bcc} \\
\midrule
\multicolumn{5}{l}{\textit{ACE}} \\
\hspace{6pt}Qwen 27B (4) & 69.0 & 56.6 & 80.4 & \cellcolor{black!10}\textbf{80.6} \\
\hspace{6pt}Qwen 9B (3) & 27.8 & 25.5 & 41.4 & \cellcolor{black!10}\textbf{43.3} \\
\hspace{6pt}Gemma 31B (4) & 59.7 & 65.1 & 70.3 & \cellcolor{black!10}\textbf{71.7} \\
\multicolumn{5}{l}{\textit{Dynamic Cheatsheet}} \\
\hspace{6pt}Qwen 27B (3) & 65.9 & 78.0 & \textbf{81.6} & \cellcolor{black!10}80.4 \\
\hspace{6pt}Qwen 9B (2) & 31.8 & 30.6 & \textbf{36.1} & \cellcolor{black!10}\textbf{36.1} \\
\hspace{6pt}Gemma 31B (3) & 56.3 & 65.2 & \textbf{72.8} & \cellcolor{black!10}\textbf{72.8} \\
\bottomrule
\end{tabular}}
\caption{Task success (0--100) per model, averaged over its settings and evaluation seeds; the whole playbook is averaged over the three growth runs. Best per row in bold.}
\label{tab:app-models}
\end{table}

Table~\ref{tab:app-models} averages Table~\ref{tab:main} per model, the evidence for Section~\ref{sec:main}'s
claim on every model.

\subsection{Judges given an advantage}
\label{app:strong}
\begin{table}[!ht]
\centering\small\setlength{\tabcolsep}{5pt}
\begin{tabular}{lrrrr}
\toprule
\hb{Judge} & \hd{Settings} & \hd{Score} & \hd{\cellcolor{black!10}\mCut} & \hd{Ahead} \\
\midrule
\mCountersT & 9 & 46.1 & \cellcolor{black!10}\textbf{59.7} & 6 \\
\mTopkT & 9 & 51.8 & \cellcolor{black!10}\textbf{59.7} & 7 \\
\mDCs & 4 & 33.9 & \cellcolor{black!10}\textbf{48.3} & 3 \\
\mDCo & 3 & 26.6 & \cellcolor{black!10}\textbf{37.4} & 3 \\
\bottomrule
\end{tabular}
\caption{Judges given an advantage: average task success (0--100) over the settings where each ran, beside \mCut{} on the same growth run and settings; Ahead: settings where \mCut{} scores higher. \mCountersT, \mTopkT: tuned on the evaluation tasks; \mDCs, \mDCo: an LLM curator that judges alone or is told the task outcome (Appendix~\ref{app:baselines}).}
\label{tab:strong}
\end{table}

\begin{table}[!ht]
\centering\small\setlength{\tabcolsep}{3.2pt}\renewcommand{\arraystretch}{1.05}
\fittable{12}{%
\begin{tabular}{lrrrrr}
\toprule
\hb{Setting} & \hd{\mCountersT} & \hd{\mTopkT} & \hd{\mDCs} & \hd{\mDCo} & \hd{\cellcolor{black!10}\mCut} \\
\midrule
27B AW p/f & 25.6 & 67.1 & 27.8 & -- & \cellcolor{black!10}\textbf{81.0} \\
9B WS & 31.0 & 31.5 & \textbf{51.9} & 30.2 & \cellcolor{black!10}45.9 \\
9B AW p/f & \textbf{41.3} & 24.0 & 21.4 & 17.5 & \cellcolor{black!10}22.2 \\
Gemma AW p/f & 16.9 & 67.1 & -- & -- & \cellcolor{black!10}\textbf{69.2} \\
27B WS & 57.6 & 45.7 & -- & -- & \cellcolor{black!10}\textbf{63.0} \\
9B AW p-t & 25.8 & 32.9 & 34.3 & 32.1 & \cellcolor{black!10}\textbf{44.0} \\
Gemma WS & 56.8 & 46.2 & -- & -- & \cellcolor{black!10}\textbf{58.1} \\
Gemma ALF & \textbf{86.4} & 79.9 & -- & -- & \cellcolor{black!10}83.1 \\
Gemma AW p-t & \textbf{73.8} & 72.2 & -- & -- & \cellcolor{black!10}71.2 \\
\bottomrule
\end{tabular}}
\caption{The judges of Table~\ref{tab:strong} per setting (task success, 0--100, at 40\% of the tokens; --: not run; best in bold). AW: AppWorld (p/f pass/fail, p-t per-test), WS: WebShop, ALF: ALFWorld.}
\label{tab:app-strong}
\end{table}

\begin{table}[!h]
\centering\small
\fittable{8}{%
\begin{tabular}{lrrrr}
\toprule
\hb{Qwen3.8-27B} & \hd{\mNone} & \hd{Whole} & \hd{DG-Mem} & \hd{\cellcolor{black!10}\mCut} \\
\midrule
AppWorld p/f & 78.4 & 31.2 & 72.8 & \cellcolor{black!10}\textbf{81.0} \\
AppWorld p-t & 78.4 & 82.5 & 83.5 & \cellcolor{black!10}\textbf{84.3} \\
WebShop & 44.4 & 59.0 & \textbf{64.8} & \cellcolor{black!10}63.0 \\
ALFWorld & 74.9 & 97.5 & 91.3 & \cellcolor{black!10}\textbf{95.3} \\
\bottomrule
\end{tabular}}
\caption{DG-Mem's subset-Shapley estimates in place of \bcc's regression (same practice episodes and serving rule), on the first growth run of the four 27B settings (task success, 0--100; better curated playbook in bold).}
\label{tab:app-dgmem}
\end{table}

An advantage does not make judging catch up with measuring (Table~\ref{tab:strong}; per setting in
Table~\ref{tab:app-strong}, definitions in Appendix~\ref{app:baselines}). Tuned on the evaluation tasks, or told each
task's outcome, the judges still score below \mCut{} on the same runs and settings. The estimator matters as well:
from the same practice episodes and with the same serving rule, DG-Mem's subset-Shapley estimates give a worse
curated playbook than \bcc's regression in three of the four 27B settings (Table~\ref{tab:app-dgmem}). Judging can
even undo a good choice: cutting the playbook that \mBestFull{} chose to 40\% of its tokens by similarity averages
\numMainBfTopkAce{} on ACE and \numMainBfTopkDc{} on Dynamic Cheatsheet, against \numKAvgBestFullThree{} and
\numSysDcBestfull{} for \mBestFull{} alone.

\subsection{Choosing among growth runs}
\label{app:selection}
\begin{table}[!h]
\centering\small
\fittable{8}{%
\begin{tabular}{lrrr}
\toprule
\hb{Method} & \hd{ACE} & \hd{DC} & \hd{All} \\
\midrule
Curate one run (\mCut) & 61.9 & 61.1 & 61.6 \\
Choose on one run (\mBestFull, one run) & 62.2 & 62.0 & 62.1 \\
Both on one run (\bcc, one run) & 63.1 & 62.7 & 62.9 \\
Choose among three runs (\mBestFull) & 66.1 & 66.9 & 66.4 \\
\cellcolor{black!10}Both among three runs (\bcc) & \cellcolor{black!10}67.2 & \cellcolor{black!10}66.5 & \cellcolor{black!10}66.9 \\
\bottomrule
\end{tabular}}
\caption{What each part of \bcc{} adds: average task success (0--100) over the settings of each memory system and over the three growth runs (one-run methods) or with the choice among them (three-run methods).}
\label{tab:app-parts}
\end{table}

Table~\ref{tab:app-runs} gives every growth run behind Figure~\ref{fig:runs} and Section~\ref{sec:experiments}.
Within one setting, whole playbooks and even their curated sets can lie on both sides of no playbook, which is why
\bcc{} lets practice tasks choose among them instead of trusting any single run. Table~\ref{tab:app-parts} shows what
each part of \bcc{} adds on both systems.

\begin{table}[!ht]
\centering\footnotesize\setlength{\tabcolsep}{5pt}
\fittable{12}{%
\begin{tabular}{lrrrrr}
\toprule
\hbm{Method} & \hd{27B} & \hd{9B} & \hd{9B} & \hd{27B} & \hdm{Mean} \\
 & \hd{AW p/f} & \hd{AW p/f} & \hd{AW p-t} & \hd{WS} & \\
\midrule
\mNone & 64.0 & 8.6 & 8.6 & 43.8 & 31.3 \\
\mBestFull & 64.0 & 31.6 & 24.2 & 60.9 & 45.2 \\
\cellcolor{black!10}Curated, verified & \cellcolor{black!10}64.8 & \cellcolor{black!10}32.1 & \cellcolor{black!10}28.8 & \cellcolor{black!10}64.0 & \cellcolor{black!10}47.4 \\
\midrule
Pooled, 40\% & 70.5 & 7.9 & 23.5 & 62.6 & 41.1 \\
Pooled, uncapped & 73.1 & 18.5 & 16.1 & 65.4 & 43.3 \\
\bottomrule
\end{tabular}
}
\caption{Pooling the lessons of the three growth runs instead of choosing one (pilot, four development settings; task success, 0--100, on task pools never used to choose). Pooled, 40\%: at 40\% of the mean single-run tokens. AW: AppWorld, WS: WebShop.}
\label{tab:app-pooling}
\end{table}

\tpar{Why choose rather than pool.}A natural alternative pools the lessons of the $K$ runs, merges near-duplicates,
clusters them into $8K$ blocks and curates once, with each block included with probability $1/K$
(Table~\ref{tab:app-pooling}). It scores below \mBestFull{} on average, and below no playbook on 9B AppWorld pass/fail
although the good run was in the pool: a playbook's value can lie in its lessons together, which choosing among runs
keeps and pooling breaks.

\subsection{Cost}
\label{app:cost}
\begin{table}[!ht]
\centering\small\setlength{\tabcolsep}{3pt}
\fittable{6}{%
\begin{tabular}{@{}lrrr@{}}
\toprule
\hb{Episodes per run} & \hd{AppWorld} & \hd{WebShop} & \hdR{ALFWorld} \\
\midrule
Growth (ACE) & 450 & 1{,}288 & 2{,}500 \\
Practice round & 684 & 600 & 600 \\
Practice score per candidate & 171 & 150 & 140 \\
\midrule
\multicolumn{4}{@{}l}{Total, relative to \mCut{}} \\
\bcc{} ($K = 1$) & 1.5$\times$ & 1.2$\times$ & 1.1$\times$ \\
\bcc{} ($K = 3$) & 4.1$\times$ & 3.6$\times$ & 3.3$\times$ \\
\mBestFull{} ($K = 3$) & 1.8$\times$ & 2.4$\times$ & 2.6$\times$ \\
\bottomrule
\end{tabular}}
\caption{Cost in agent episodes, per growth run and in total relative to \mCut{} (one growth run and one practice round). \bcc{} repeats both for each of its $K$ runs and scores $2K + 1$ candidates; \mBestFull{} grows $K$ runs and scores $K + 1$. Evaluation episodes are not part of any method.}
\label{tab:cost}
\end{table}

\begin{table}[!h]
\centering\small
\fittable{6}{%
\begin{tabular}{lrrr}
\toprule
\hb{Method} & \hd{Curation (M)} & \hd{Serving (k)} & \hd{Success} \\
\midrule
\mNone & 0 & 59 & 54.4 \\
\mFull & 227 & 212 & 51.2 \\
\mCut & 321 & 115 & 61.9 \\
\mBestFull, 1 run & 278 & 161 & 62.2 \\
\mBestFull, 2 runs & 546 & 159 & 64.3 \\
\mBestFull, 3 runs & 814 & 166 & 66.1 \\
\mBestFull, 4 runs & 1,089 & 175 & 65.9 \\
\mBestFull, 5 runs & 1,359 & 175 & 66.8 \\
\cellcolor{black!10}\bcc, 1 run & \cellcolor{black!10}394 & \cellcolor{black!10}141 & \cellcolor{black!10}63.1 \\
\cellcolor{black!10}\bcc, 2 runs & \cellcolor{black!10}778 & \cellcolor{black!10}135 & \cellcolor{black!10}65.7 \\
\cellcolor{black!10}\bcc, 3 runs & \cellcolor{black!10}1,162 & \cellcolor{black!10}138 & \cellcolor{black!10}67.2 \\
\bottomrule
\end{tabular}}
\caption{Cost in tokens on ACE, mean over the eleven settings: millions spent on growth and practice episodes, thousands per served task, and task success (0--100). \mBestFull{} with four and five runs grows two more runs per setting.}
\label{tab:app-tokens}
\end{table}

\bcc{} spends its cost before serving, and saves at serving time. Curating one run adds one practice round to
growth, in which every episode measures all blocks at once (Table~\ref{tab:cost}); \bcc{} repeats growth and the
round for each of its $K$ runs and scores its $2K + 1$ candidates, \numCostRatioMin--\numCostRatioMax{} times the
episodes of curating one run at $K = 3$, against \numCostRatioBestFullMin--\numCostRatioBestFullMax{} times for
\mBestFull. In tokens (Table~\ref{tab:app-tokens}), which also count the Reflector and Curator calls of growth, \bcc{}
with three runs serves fewer tokens per task than \mBestFull{} with any number of runs, and \mBestFull{} needs two more
runs, and \numAkCurationPct\% more growth and practice tokens, to come level in success.

\subsection{Long growth}
\label{app:long}
Long growth changes two choices of the growth recipe, both fixed before any result: up to ten passes over the
training tasks instead of five, and no early stop. Everything else, including \bcc{} and \mBestFull, is unchanged,
and each setting gets three new growth runs.

\subsection{Checking on the evaluation distribution}
\label{app:crossfit}
\begin{table}[!h]
\centering\footnotesize
\fittable{6}{%
\begin{tabular}{@{}lrrr@{}}
\toprule
\hb{Gain of \bcc{}} & \hd{ACE} & \hd{DC} & \hdR{All} \\
\midrule
\multicolumn{4}{@{}l}{\textit{Checked on practice tasks (the method)}} \\
\hspace{6pt}minus \mNone & \ensuremath{+}12.8 & \ensuremath{+}12.7 & \ensuremath{+}12.8 \\
\hspace{6pt}minus \mFull & \ensuremath{+}16.0 & \ensuremath{+}5.2 & \ensuremath{+}11.4 \\
\hspace{6pt}minus \mBestFull & \ensuremath{+}1.1 & \ensuremath{-}0.4 & \ensuremath{+}0.5 \\
\addlinespace[2pt]
\multicolumn{4}{@{}l}{\textit{Checked on evaluation halves}} \\
\hspace{6pt}minus \mNone & \ensuremath{+}13.2 & \ensuremath{+}14.9 & \ensuremath{+}13.9 \\
\hspace{6pt}minus \mFull & \ensuremath{+}16.4 & \ensuremath{+}7.4 & \ensuremath{+}12.6 \\
\hspace{6pt}minus \mBestFull & \ensuremath{+}1.0 & \ensuremath{+}0.3 & \ensuremath{+}0.7 \\
\bottomrule
\end{tabular}}
\caption{Gain of \bcc{} (points, averaged over tasks) when the choice among candidates is checked on practice tasks, as in the method, or on one half of the evaluation tasks and scored on the other, averaged over 50 random halvings; \mBestFull{} is checked the same way.}
\label{tab:app-crossfit}
\end{table}

\begin{figure}[!h]
\centering
\includegraphics[width=\columnwidth]{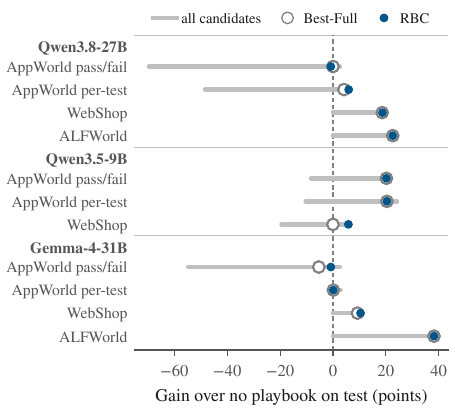}
\caption{Gain over no playbook on the evaluation tasks for the seven candidates of each setting (grey: worst to best) and the candidates that \bcc{} (blue) and \mBestFull{} (open) pick on practice tasks.}
\label{fig:transfer}
\end{figure}

AppWorld's practice tasks come from its development split and its evaluation tasks from \texttt{test\_normal}
(Table~\ref{tab:cells}). To test whether this difference causes the misses of Section~\ref{sec:selection}, we
checked the candidates on tasks of the evaluation split without choosing on the tasks that are scored: each
setting's evaluation tasks are split in half, the candidate is chosen on one half and scored on the other, and the
halves are swapped. We average over \numSysCrossR{} random halvings and check \mBestFull{} in the same way.
Table~\ref{tab:app-crossfit} compares the two checks. Every conclusion of Section~\ref{sec:experiments} holds under
both. Checked on the evaluation split, \bcc{} gains more over every reference and moves ahead of \mBestFull{} on
Dynamic Cheatsheet as well, and the AppWorld settings that fell below no playbook reach it or come close: the
practice split, not the method, explains most of the misses.

\subsection{Robustness}
\label{app:sensitivity}
\begin{table}[!ht]
\centering\small\setlength{\tabcolsep}{5pt}
\begin{tabular}{@{}lr@{\hspace{14pt}}lr@{}}
\toprule
\hb{Change} & \multicolumn{1}{c@{\hspace{14pt}}}{\hb{Score}} & \hb{Change} & \hdR{Score} \\
\midrule
Default & 47.4 & $m = 4$ & 45.9 \\
$K = 4$ & 48.1 & $m = 16$ & 48.0 \\
$K = 5$ & 48.1 & $B = 20\%$ & 46.1 \\
Choice on $2\times$ episodes & 47.4 & $B = 60\%$ & 46.1 \\
\bottomrule
\end{tabular}
\caption{Tuning round on the four development settings: mean task success (0--100) on held-out task pools when one choice at a time moves from the default ($K = 3$ runs, $B = 40\%$, $m = 8$ blocks). \mBestFull{} scores 45.2 in every row of the table.}
\label{tab:app-tune}
\end{table}

\tpar{Development and tuning.}We chose $K = 3$ in a pilot on the four development settings, scored on held-out task
pools (AppWorld's \texttt{test\_challenge} split and half of WebShop's practice tasks), never on evaluation tasks. A tuning round on the same settings then moved one choice at a time from this default ($K = 3$, $B = 40\%$,
$m = 8$; Table~\ref{tab:app-tune}), under a rule fixed in advance: a choice moves only if it gains at least one point
on the four-setting mean. None did, so the default ran unchanged on every setting.

\begin{table}[!ht]
\centering\footnotesize\setlength{\tabcolsep}{3pt}
\fittable{12}{%
\begin{tabular}{lrrrrr}
\toprule
\hbm{Setting} & \hdm{\mNone} & \hdm{\mFull} & \hg{2}{\mCut} & \hd{Tokens} \\
\cmidrule(lr){4-5}
 & & & \hs{\cellcolor{black!10}40\%} & \hs{uncapped} & \hd{(\%)} \\
\midrule
27B AW p/f & 78.4 & 31.2 & \cellcolor{black!10}\textbf{81.0} & \textbf{81.0} & 8 \\
9B WS & 43.8 & 24.3 & \cellcolor{black!10}\textbf{45.9} & \textbf{45.9} & 38 \\
9B AW p/f & 19.8 & 13.5 & \cellcolor{black!10}\textbf{22.2} & \textbf{22.2} & 44 \\
Gemma AW p/f & \textbf{70.6} & 15.9 & \cellcolor{black!10}69.2 & 68.8 & 47 \\
\midrule
27B WS & 44.4 & 59.0 & \cellcolor{black!10}\textbf{63.0} & 61.1 & 56 \\
27B ALF & 74.9 & \textbf{97.5} & \cellcolor{black!10}95.3 & 95.3 & 66 \\
27B AW p-t & 78.4 & 82.5 & \cellcolor{black!10}\textbf{84.3} & 83.5 & 75 \\
9B AW p-t & 19.8 & 40.3 & \cellcolor{black!10}\textbf{44.0} & 38.9 & 82 \\
Gemma WS & 47.7 & 57.0 & \cellcolor{black!10}58.1 & \textbf{58.9} & 73 \\
Gemma ALF & 50.6 & \textbf{84.3} & \cellcolor{black!10}83.1 & 75.9 & 67 \\
Gemma AW p-t & 69.9 & 70.1 & \cellcolor{black!10}\textbf{71.2} & 70.6 & 72 \\
\midrule
Mean & 54.4 & 52.3 & \cellcolor{black!10}\textbf{65.2} & 63.8 & 57 \\
\bottomrule
\end{tabular}
}
\caption{\mCut{} with and without the 40\% cap, from the same estimates (task success, 0--100; best per row in bold). Uncapped serves every block with a positive estimate; Tokens: its share of the whole playbook. AW: AppWorld (p/f pass/fail, p-t per-test feedback), WS: WebShop, ALF: ALFWorld.}
\label{tab:app-cap}
\end{table}

\tpar{The cap is part of the method.}Serving every block with a positive estimate, instead of the best blocks within
40\% of the tokens, scores lower from the same estimates (\numVthreeFreeAvgDone{} against \numVthreeAvgDone;
Table~\ref{tab:app-cap}). The loss sits on Gemma ALFWorld and 9B AppWorld per-test. On Gemma ALFWorld, practice and
evaluation estimates agree, so the loss comes from blocks with small positive effects that do not add up when served
together, not from noise; the cap keeps only the blocks whose effect is clear.

\begin{table}[!ht]
\centering\footnotesize\setlength{\tabcolsep}{3pt}
\fittable{14}{%
\begin{tabular}{@{}lrrrrrrr@{}}
\toprule
\hb{Round size ($\times$ as run)} & \hd{$\frac14$} & \hd{$\frac12$} & \hd{$\frac34$} & \hd{1} & \hd{2} & \hd{4} & \hdR{$\infty$} \\
\midrule
Subsampled rounds & 8.9 & 10.2 & 11.0 & 10.8 & & & \\
Variance model & & & & 9.8 & 10.6 & 11.3 & 12.6 \\
\bottomrule
\end{tabular}
}
\caption{Gain of \bcc's 40\% set over no playbook by practice-round size (points, out of sample, mean of 9 settings). Top: the round subsampled; bottom: a model of the estimates' variance ($\infty$: perfect measurement).}
\label{tab:app-coins}
\end{table}

\tpar{One practice round is enough.}When the round is subsampled, the value of the curated set rises up to three
quarters of the round and is flat from there (Table~\ref{tab:app-coins}); a model of the estimates' variance predicts
that a round two or four times larger would add less than the evaluation noise of the comparisons.

\subsection{Controls}
\label{app:controls}
\begin{table}[!ht]
\centering\footnotesize\setlength{\tabcolsep}{3pt}
\fittable{6}{%
\begin{tabular}{lrr}
\toprule
\hb{9B WebShop} & \hd{Thinking off} & \hd{Thinking on} \\
\midrule
\mNone & 43.8 & 46.3 \\
\mFull & 24.3 & 26.4 \\
\midrule
\mRandom & 23.8 & 30.3 \\
\mCountersT & 31.0 & 35.4 \\
\mTopkT & 31.5 & 33.0 \\
\cellcolor{black!10}\bcc{} (earlier rule) & \cellcolor{black!10}\textbf{43.7} & \cellcolor{black!10}\textbf{39.6} \\
\bottomrule
\end{tabular}
}
\caption{Thinking on in every role (task success, 0--100). Below the rule: 40\% of the tokens, best in bold. Run with the earlier serving rule; growth with thinking is capped at two passes, so each column is compared with its own \mNone{} and \mFull.}
\label{tab:app-thinking}
\end{table}

\tpar{The harm is not an artefact of thinking off.}We regrew the 9B WebShop playbook with thinking on in every role. With thinking on as with it off, the whole playbook scores below no playbook and the
curated set above every tag and retrieval rule (Table~\ref{tab:app-thinking}).

\tpar{Task success cannot be replaced by an LLM judge.}On AppWorld we also estimated block effects from a goal check
by the model itself instead of task success. With every judge prompt we tried, including a judge from another model, the set curated on the judge's estimates failed on 27B and 9B
AppWorld pass/fail, in line with the known biases of LLM judges \citep{zheng2023judging, wang2024fair}: the measured
outcome is what makes \bcc{} work.

\section{Comparison with related methods}
\label{app:related}

Table~\ref{tab:related} places \bcc{} beside the closest methods on the axes of Section~\ref{sec:related}.
\begin{table}[!ht]
\centering\footnotesize\setlength{\tabcolsep}{3.5pt}
\fittable{10}{%
\begin{tabular}{lcccc}
\toprule
\hb{Method} & \hd{Success} & \hd{All units} & \hd{Set} & \hd{Runs} \\
\midrule
\rowcolor{black!10}\bcc{} (ours) & \yes & \yes & \yes & \yes \\
SkillSV & \half & \no & \no & \no \\
DG-Mem & \half & \half & \no & \no \\
SkillShapley & \half & \no & \no & \no \\
Coalition-aware & \no & \half & \no & \no \\
ContextCite & \no & \yes & \no & \no \\
\mCounters & \half & \yes & \half & \no \\
ExpeL & \half & \no & \half & \no \\
RSEA & \yes & \no & \half & \half \\
\bottomrule
\end{tabular}}
\caption{\bcc{} beside the closest prior methods (\yes{} yes, \half{} in part, \no{} no). Success: the response
is task success (\half: a verifier, a gold answer or the agent's own judgement). All units: one run informs every
unit, not one marginal. Set: outputs a set to serve under a budget, not a value per unit. Runs: chooses among
growth runs.}
\label{tab:related}
\end{table}

\tpar{ContextCite \citep{cohenwang2024contextcite}.}Masks the sources of one context at random and fits a sparse
linear model of the log-probability of the original answer on the masks. Its design, random inclusion and a linear
fit, is the one we use; its response, the chance of reproducing an earlier output rather than task success, is not.

\tpar{SkillSV \citep{skillsv2026}.}A truncated permutation Shapley value over the units of an optimiser-written
skill file, each addition scored by a verifier, so each rollout gives one marginal for one unit.

\tpar{DG-Mem, SkillShapley and coalition-aware valuation.}DG-Mem \citep{dgmem2026} values memory rules on random
subsets of at most four rules against gold answers; SkillShapley \citep{skillshapley2026} enumerates skill
configurations; coalition-aware valuation \citep{coalitionaware2026} masks groups of skills and scores them by
retrieval similarity, and reports that single-skill knockouts resolve only small effects at a few hundred queries,
the problem our blocks answer. All three output a value per unit, not a set to serve.

\tpar{ACE's tags and memory systems.}ACE's Reflector tags every lesson an episode used, but the response is its own
judgement, and ACE never removes a lesson \citep{zhang2025ace}; \mCounters{} is the rule its tags
imply (Appendix~\ref{app:baselines}). ReasoningBank \citep{ouyang2025reasoningbank} serves the items most similar to
the query (\mTopk), and Dynamic Cheatsheet \citep{suzgun2025dynamic} lets the model rewrite its memory and count
each entry's successful uses; \mDCs{} and \mDCo{} run its curator as a judge. None ranks by a measured effect on success, and none chooses between growth runs.

\tpar{ExpeL and RSEA.}ExpeL \citep{zhao2024expel} keeps an importance count per insight, raised and lowered by the
model's own votes, and drops an insight at zero; its signal is the agent's judgement, as with ACE's tags. RSEA
\citep{rsea2026} accepts a rewritten memory only if held-out success does not drop: a measured outcome, but for the
memory as a whole and for one run. \mBestFull{} applies its rule to one and to three growth runs
(Appendix~\ref{app:baselines}).

\section{Prompts}
\label{app:prompts}
\bcc{} calls no language model: it serves blocks of lessons at random and fits a regression on task success. Every
prompt in our runs therefore belongs to the actor or to the memory system (Table~\ref{tab:prompts}). We wrote the
WebShop and ALFWorld actor prompts; all other prompts are published, and we added one line to one of them.

\begin{table}[!ht]
\centering\small
\begin{tabular}{@{}l>{\raggedright\arraybackslash}p{4.05cm}@{}}
\toprule
\hb{Prompt} & \hb{Source} \\
\midrule
Actor, AppWorld & AppWorld's ReAct prompt in ACE's code, with the playbook block added \\
Actor, WebShop & ours \\
Actor, ALFWorld & ours \\
ACE Reflector & ACE, without ground truth, with one line added \\
ACE Curator & ACE, without ground truth \\
ACE aggregation, merge & ACE \\
DC curator & Dynamic Cheatsheet \\
\bottomrule
\end{tabular}
\caption{The prompts of our runs and where they come from. \bcc{} adds none.}
\label{tab:prompts}
\end{table}

\tpar{The playbook block.}Every actor sees the playbook under one fixed heading, one lesson per line, without the
tags. This block is the only part of the prompt that differs between methods.

\tpar{WebShop and ALFWorld.}Both actors are zero-shot ReAct loops whose system message holds the instructions below
and the playbook block.

\begin{promptbox}{WebShop actor, system message}
You are shopping on a web store for the product described in the instruction. Each turn, reply with exactly one
action on a single line and nothing else. Actions: search[\textless keywords\textgreater] (only on the search page),
and click[\textless item\textgreater] where \textless item\textgreater{} is any text shown in square brackets on
the current page (a product id, an option value, Back to Search, Next \textgreater, \textless{} Prev, Description,
Features, Reviews, Buy Now). Clicking Buy Now purchases the product on the current page and ends the task. You may
also reply ``think[\textless your reasoning\textgreater]'' to plan; thinking does not change the page. You have at
most 15 actions, thinking included.\par\smallskip
\#\# Playbook (strategies learned from past tasks)\\
- [\textit{id}] \textit{lesson}, one line per lesson
\end{promptbox}

\begin{promptbox}{ALFWorld actor, system message}
Interact with a household to solve a task. Each turn, reply with exactly one action on a single line and nothing
else. Valid actions include: go to \textless recep\textgreater, open \textless recep\textgreater, close
\textless recep\textgreater, take \textless obj\textgreater{} from \textless recep\textgreater, move
\textless obj\textgreater{} to \textless recep\textgreater, heat \textless obj\textgreater{} with
\textless recep\textgreater, cool \textless obj\textgreater{} with \textless recep\textgreater, clean
\textless obj\textgreater{} with \textless recep\textgreater, use \textless obj\textgreater, examine
\textless obj\textbar recep\textgreater, look, inventory. You may also reply ``think: \textless your
reasoning\textgreater'' to plan; thinking does not change the environment. The task ends automatically when it is
completed; never reply ``done'' or similar.\par\smallskip
\#\# Playbook (strategies learned from past tasks)\\
- [\textit{id}] \textit{lesson}, one line per lesson
\end{promptbox}

\tpar{AppWorld.}The actor uses the AppWorld prompt from ACE's code \citep{zhang2025ace, trivedi2024appworld},
with the playbook block added after the instructions.

\tpar{ACE's Reflector.}Our actors do not cite lessons, so the Reflector sees the whole playbook and is told that
lessons it does not tag count as neutral. All other memory prompts run unchanged.

\end{document}